\documentclass[a4paper,fleqn]{cas-dc}

\usepackage[numbers]{natbib}
\usepackage[ruled,lined,longend,linesnumbered]{algorithm2e}

\usepackage{algpseudocode}
\usepackage{booktabs}
\usepackage{multirow}
\usepackage{amsmath} 
\usepackage{amssymb}   
\usepackage{mathtools} 
\usepackage{amsthm}

\usepackage{cite}
\usepackage{subcaption}
\usepackage{multirow}
\usepackage{caption}
\usepackage{hyperref}
\usepackage{microtype}
\usepackage{tabularx}
\usepackage{placeins}
\usepackage{makecell}
\usepackage{hyperref}
\usepackage{orcidlink}

\def\tsc#1{\csdef{#1}{\textsc{\lowercase{#1}}\xspace}}
\tsc{QE}
\tsc{EP}
\tsc{PMS}
\tsc{DE}

\newtheorem{theorem}{Theorem}

\shorttitle{FedA2L: Adaptive Layer-wise Learning Rate Adjustment in Decentralized Federated Learning}
\shortauthors{Truong et~al.}

\title[mode = title]{FedA2L: Adaptive Layer-wise Learning Rate Adjustment in Decentralized Federated Learning}

\author{Van Truong Vo\orcidlink{0009-0004-5200-0640}}
\ead{truongvo@cbnu.ac.kr}

\author{Khoa Nguyen\orcidlink{0009-0006-8840-8140}}
\ead{neko941@cbnu.ac.kr}

\author{Taehong Kim\corref{cor1}\orcidlink{0000-0001-6246-6218}}
\ead{taehongkim@cbnu.ac.kr}
\cortext[cor1]{Corresponding author}

\nonumnote{%
\textcopyright{} 2026. This manuscript version is made available under the
CC-BY-NC-ND 4.0 license
(\url{https://creativecommons.org/licenses/by-nc-nd/4.0/}).
This is the accepted manuscript version of an article published in
\textit{Future Generation Computer Systems}. The final published version
is available at
\url{https://doi.org/10.1016/j.future.2026.108743}.
}

\affiliation{
    organization={School of Information and Communication Engineering, Chungbuk National University},
    city={Cheongju},
    country={South Korea}
}
\begin{abstract}
Decentralized intelligence systems with heterogeneous devices and limited coordination increasingly rely on decentralized federated learning (DFL). However, DFL suffers from convergence inefficiency under data heterogeneity due to the use of a uniform learning rate (LR) that ignores layer-specific optimization needs. Foundational layers are responsible for maintaining network consensus, while specialized layers adapt to local data characteristics, leading to conflicting gradients and degraded performance under non-IID conditions. To address this fundamental tension, this work introduces FedA2L, a method that dynamically adjusts layer-wise LRs based on model divergence signals. By leveraging local update intensity and network consensus constraints, FedA2L seamlessly integrates into existing DFL protocols without additional communication or coordination. Extensive evaluations across DFL algorithms, various model architectures, and datasets demonstrate that FedA2L achieves up to 4.94$\times$ faster convergence than vanilla DFL and reduces communication rounds by up to 59\% compared to scheduler-based baselines. Furthermore, FedA2L exhibits resilience to severe data heterogeneity, larger network sizes, and sparse topologies, reducing communication overhead and establishing it as a versatile optimization tool for resource-constrained or large-scale distributed learning in edge and IoT deployments. 
The code is released at \href{https://github.com/nclabteam/FedA2L}{https://github.com/nclabteam/FedA2L}.

\end{abstract}

\begin{document}


\maketitle

\section{Introduction}
\label{Intro}

Recent advances in Industrial IoT (IIoT) and Cyber-Physical Systems (CPS) require robust decentralized intelligence mechanisms to maintain real-time synchronization between physical assets and their Digital Twins (DT) \citep{9244624}. In such settings, centralized cloud-based learning introduces unacceptable latency, privacy risks, and single points of failure \citep{padmavathi2025digital}. In this context, federated learning (FL) has emerged as a distributed computing paradigm that enables collaborative model training on decentralized data, preventing direct access to sensitive information at individual nodes \citep{mcmahan2017communication}. In standard FL, a central server coordinates training by collecting local model updates from participating clients, aggregating them into a global model, and broadcasting updated parameters back to clients for the next round. However, this centralization introduces vulnerabilities, such as communication bottlenecks \citep{wang2021edge}, and scalability issues \citep{ wang2025verifydfl,yuan2024decentralized}. 

DFL mitigates these issues by removing the central server, enabling direct peer-to-peer (P2P) communication among nodes according to a network topology \citep{lalitha2018fully}. In DFL, each node independently performs local model training on its private data and exchanges model parameters only with direct neighbors, then aggregates received models to achieve network consensus without central coordination \citep{wang2023distributed}. This architecture enhances robustness, alleviates communication bottlenecks by distributing aggregation load across the network, and improves scalability. These properties make DFL well suited for large-scale networked systems, including edge intelligence systems~\citep{10839118,chen2024resilient}, Internet of Things (IoT) networks~\citep{11134848,molo2025decentralized}, and traffic-aware communication environments~\citep{10704927}, where centralized coordination is often impractical.

Despite these advantages, DFL systems confront a critical challenge of statistical heterogeneity in distributed data. In practice, data on each node is non-identically and independently distributed (non-IID). This heterogeneity causes each node's local learning objectives to drift away from the consensus objectives enforced by neighbor-wise aggregation, a phenomenon known as client drift \citep{li2020federated,selo2025fedtvd}. This drift manifests as inconsistency between local gradients and the global optimization direction, leading to conflicting parameter updates. These conflicts slow convergence and reduce model quality \citep{gao2022feddc,park2025def}. This problem intensifies as data heterogeneity becomes more severe, creating a critical obstacle to achieving efficient and accurate distributed learning. 

This problem is further compounded in deep neural networks (DNNs), where client drift emerges unevenly across model layers owing to their hierarchical structure in feature extraction. Foundational layers capture general and shared features to support network consensus, while specialized layers adapt to task-specific features in local data \citep{yosinski2014transferable, zeiler2014visualizing}. Within DFL, this architectural hierarchy creates fundamentally different optimization challenges for each layer. Specialized layers are highly adapted to local data characteristics to achieve good performance on local objectives, causing significant dispersion between nodes. Conversely, foundational layers that over-homogenize shared representations limit their capacity to maintain task-general features to support local performance \citep{elhussein2025playerfl}. Recent work on layer-wise personalized FL \citep{chen2024optimizing} confirms this asymmetry, demonstrating that optimal layer-wise treatment requires balancing local update intensity in specialized layers with network consensus constraints in foundational layers. 

This tension is exacerbated by the widespread use of a uniform LR across all model layers in standard DFL algorithms. Treating the LR as a single scalar hyperparameter forces all layers to update at the same rate, ignoring the distinct optimization landscapes, gradient magnitudes, and update sensitivities exhibited by different layers. As a result, uniform LRs in DFL systems suffer from training instability, prolonged convergence time, and suboptimal model quality, becoming more significant in the case of severe non-IID conditions.

To address this critical gap in DFL systems, this study proposes the ``\textbf{A}daptive \textbf{L}ayer-wise \textbf{L}earning Rate Adjustment in Decentralized \textbf{Fed}erated Learning'' (FedA2L), a serverless method for dynamically tuning layer-wise LRs based on model divergence signals. The core innovation is formalizing model states at distinct transitions of each DFL round: the base state (before local training), the trained state (after local gradient updates), and the aggregated state (after network consensus). Through analyzing transitions between these three states, FedA2L derives two layer-wise metrics capturing distinct aspects of the optimization dynamics: weight divergence ($\sigma$) measuring local update intensity across nodes and aggregation stability ($\zeta$) measuring network consensus constraints. By leveraging these signals, FedA2L dynamically assigns a unique LR to each model layer, enabling simultaneous optimization of local learning and network consensus without requiring central coordination or additional communication overhead. The primary contributions can be summarized as follows:

\begin{itemize}
    \item FedA2L introduces a state-based formalization that characterizes model dynamics across three states of each DFL round, enabling precise analysis of layer-wise divergence and aggregation behavior.
    \item Adaptation of two layer-wise metrics (weight divergence $\sigma$ and aggregation stability $\zeta$) enables layer-wise dynamic LR adjustment to quantify and balance local update intensity and network consensus constraints.
    \item Layer-wise adaptive LRs based on local model divergence metrics can be seamlessly integrated into existing DFL algorithms without modifying the core aggregation logic or introducing additional communication overhead. 
    \item Extensive empirical validation demonstrates that FedA2L significantly accelerates convergence across six DFL algorithms and DFedHPO, spanning multiple neural network architectures and diverse datasets, achieving up to 4.94$\times$ faster convergence than vanilla DFL and up to $59\%$ fewer communication rounds than the best-performing scheduler-based baselines.
\end{itemize}
These contributions are particularly valuable for decentralized intelligence systems operating under dynamic conditions, where layer-wise adaptive mechanisms balance localized learning and stable network consensus across distributed entities.

The remainder of this paper is organized as follows. Section~\ref{sec:rw} reviews related work on DFL and adaptive hyperparameter optimization methods. Section~\ref{sec:method} presents the problem formulation, the three-state model of DFL node procedures, and the proposed dual-metric layer-wise LR adaptation mechanism with its algorithmic integration. To formally validate these properties, Section~\ref{sec:theory} establishes convergence guarantees for both non-convex and strongly convex settings. Section~\ref{sec:exp} evaluates convergence speed, model accuracy, robustness under severe data heterogeneity, node scalability, and sparse network topologies, together with ablation and hyperparameter sensitivity analyses of FedA2L. Section~\ref{sec:discuss} discusses the advantages and limitations of FedA2L and outlines directions for future work. Section~\ref{sec:con} concludes the paper.

\section{Related work} \label{sec:rw}
\begin{table*}[t]
\centering
\caption{Comparison of adaptive learning-rate methods for federated learning and decentralized federated learning.}
\label{tab:lr_comparison}
\resizebox{\linewidth}{!}{%
\begin{tabular}{l|c|c|c|c|l}
\toprule
Method &
Layer-wise LR &
Dynamic per Round &
Server-free &
No Extra Comm. &
Signal Source \\
\midrule

FedYogi~\citep{reddi2021adaptive}  & $\times$ & \checkmark & $\times$ & \checkmark & Server-side second moment \\
FLARE~\citep{xiao2025flare}    & $\times$ & \checkmark & $\times$ & \checkmark & Device capability \\
DFedHPO~\citep{DFedHPO}  & $\times$ & $\times$ & \checkmark & \checkmark & Pre-training search \\
AutoLR~\citep{ro2021autolr}   & \checkmark & \checkmark & $\times$ & $\times$ & Gradient norms \\
Fed-LAMB~\citep{karimi2023fed} & \checkmark & \checkmark & $\times$ & $\times$ & Gradient statistics \\
FLAYER~\citep{chen2024optimizing}   & \checkmark & \checkmark & $\times$ & $\times$ & Global gradient/loss signals \\
FedLWS~\citep{shi2025fedlws}   & \checkmark & \checkmark & $\times$ & $\times$ & Weight shrinking \\
\midrule
\textbf{FedA2L} & \checkmark & \checkmark & \checkmark & \checkmark & Local state transitions \\
\bottomrule
\end{tabular}%
}
\end{table*}

\subsection{Decentralized federated learning}
DFL eliminates the reliance on a central server, enabling robust and scalable P2P collaboration. Foundational work such as D-PSGD~\citep{lian2017can} provided convergence guarantees using P2P model averaging. Recent surveys \citep{beltran2023decentralized,yuan2024decentralized} systematically categorize DFL advances, with research focusing on enhancing communication efficiency, convergence speed under non-IID conditions, and robustness to security and privacy threats~\citep{WANG2023449,ZHOU2024120582}. This includes analyzing the impact of network topology~\citep{nedic2018network,van2025performance} and developing advanced protocols to enhance communication efficiency, such as using sparsification and quantization to compress model updates~\citep{Koloskova2019,tang2022gossipfl}, reducing transmission overhead in bandwidth-constrained edge environments. Advanced protocols, such as FedAWA~\citep{shi2025fedawa} and FedAW~\citep{tang2024adapted}, enhance aggregation by reweighing client contributions based on local data characteristics or model similarity, improving alignment with the overall learning dynamics. However, these methods assume uniform LRs during local training or apply only basic decay strategies across all layers, limiting adaptability to heterogeneous data and hierarchical model dynamics. While promising for general scenarios, such limitations hinder the deployment of robust decentralized intelligence in complex cyber-physical infrastructures that require coordinated learning across diverse and heterogeneous devices.

\subsection{Hyperparameter optimization in FL and DFL}
The LR is a critical hyperparameter in distributed optimization, affecting both convergence speed and model quality. In centralized FL, server-side adaptive optimization methods have been developed to enhance performance. FedYogi \citep{reddi2021adaptive} employs a server-side adaptive optimizer that stabilizes convergence by maintaining coordinate-wise second moment estimates. FLARE \citep{xiao2025flare} dynamically adjusts LRs for selected devices based on their individual computational capabilities. Recent work on federated hyperparameter optimization \citep{kundroo2023federated,nakka2024federated} highlights the challenge of balancing efficiency with adaptive tuning. However, all these methods are fundamentally server-dependent and incompatible with DFL. In DFL settings, DFedHPO~\citep{DFedHPO} performs a one-time, decentralized hyperparameter search prior to model training and outputs a consensus hyperparameter configuration that remains fixed during subsequent communication rounds. While these methods eliminate the need for a server, both server-dependent adaptive methods and one-shot static search are unable to accommodate 
evolving layer-specific heterogeneity or continual drift during training.

Layer-wise adaptive methods, motivated by hierarchical feature learning \citep{yosinski2014transferable,zeiler2014visualizing}, have been explored in centralized FL. AutoLR~\citep{ro2021autolr} adjusts per-layer rates based on gradient norms, while Fed-LAMB~\citep{karimi2023fed} incorporates layer-wise and dimension-wise adaptivity. More recently, FLAYER~\citep{chen2024optimizing} adjusts layer-specific LRs using gradient or loss signals, and FedLWS~\citep{shi2025fedlws} proposes adaptive layer-wise weight shrinking. However, these methods face fundamental limitations in DFL: (1) they require centralized coordination for global layer-wise statistics, (2) incur significant computational overhead, and (3) treat aggregation as a post-processing step rather than an optimization signal source. Table~\ref{tab:lr_comparison} summarizes these limitations and positions FedA2L against existing hyperparameter optimization methods across the four properties that define an effective DFL solution.

Current DFL methods often overlook the internal dynamics of local training, relying on static LRs that fail to account for node-specific data heterogeneity and layer-specific learning dynamics. Existing adaptive LR methods are either server-dependent, operate at a coarse model level, or perform static hyperparameter searches. 

To the best of our knowledge, no prior work has developed a dynamic, layer-wise adaptive LR method tailored to the serverless nature of DFL. This highlights the need for decentralized mechanisms that can respond to layer-specific optimization behavior without a centralized coordinator. To address this gap, an effective solution should rely solely on locally available information, incur no additional communication overhead, and support layer-wise adaptation to capture heterogeneous optimization roles across model layers while preserving training stability under severe non-IID conditions. 

Based on these requirements, we design a serverless, communication-efficient, and layer-wise adaptive LR mechanism for DFL. The key insight is that the standard DFL training loop already contains locally observable signals: the parameter transitions during local training and network consensus directly encode both local update intensity and network consensus constraints at the layer level. By extracting and combining these signals without any additional coordination, the proposed method achieves lightweight yet effective per-layer rate control that integrates seamlessly into existing DFL protocols, as detailed in the following section.
\section{Methodology} \label{sec:method}
\subsection{Decentralized federated learning setup} \label{sec:dfl_setup}
We consider a DFL system comprising a set of nodes $\mathcal{N}$ interconnected in a P2P topology. Each node $i \in \mathcal{N}$ holds a private local dataset $\mathcal{D}_i$, which remains strictly local and unshared. Instead of relying on a centralized server, each node $i$ collaborates by exchanging model parameters only with its direct neighbors $\mathcal{V}_i$. The objective is to learn model parameters that minimize the average of local loss functions across the network:

\begin{equation}
    \min_{\theta_1, \ldots, \theta_{|\mathcal{N}|}} \mathcal{F}(\theta_1, \ldots, \theta_{|\mathcal{N}|}) \triangleq \frac{1}{|\mathcal{N}|} \sum_{i=1}^{|\mathcal{N}|} \mathcal{L}_i(\theta_i).
    \label{eq:global_objective_full}
\end{equation}
where $\mathcal{L}_i(\theta_i) = \mathbb{E}_{(x,y) \sim \mathcal{D}_i}[\ell(\theta_i; x, y)]$, and $\ell(\cdot)$ denotes a standard loss function. Solving this optimization problem requires iterative training through communication rounds, alternating between local training on heterogeneous data and network consensus aggregation.

\subsection{Problem definition} \label{sec:problem}

Statistical heterogeneity in distributed federated systems creates uneven, layer-specific optimization challenges with significant implications for system convergence efficiency and practical deployment. During local training, specialized layers adapt rapidly to node-specific data, whereas foundational layers preserve task-general representations. Conversely, during network consensus, parameter averaging applies uniform pressure across all layers, stabilizing foundational representations but potentially suppressing necessary adaptation in specialized layers. This layer-wise disparity creates competing optimization objectives: local learning drives specialized layers toward node-specific optima, while network consensus pushes foundational layers toward network-wide agreement. 

Despite the disparity observed across layers, existing DFL methods typically apply a single scalar LR to all model layers, treating $\eta$ as a global hyperparameter that modulates updates uniformly regardless of layer position or role. This uniform approach creates a fundamental optimization conflict. Low LRs help maintain consensus in foundational layers and stabilize the network, but they limit the ability of specialized layers to adapt to local data. In contrast, high LRs accelerate adaptation in specialized layers but destabilize foundational layers, causing parameter oscillations and degraded convergence. This dilemma becomes more pronounced under severe non-IID conditions, where conflicting local optima across heterogeneous nodes further widen the gap between local update intensity and network consensus constraints. 

As a result, uniform LRs produce unstable parameter trajectories, prolonged convergence measured in communication rounds, and suboptimal model quality. In bandwidth-constrained and latency-sensitive edge deployments, resolving this conflict efficiently is critical: every communication round saved through improved convergence translates directly to reduced bandwidth consumption, lower energy expenditure, and faster system response time. Such efficiency gains are especially critical for large-scale IoT and mobile edge network scenarios where communication and energy costs directly impact system viability. Addressing this limitation requires dynamic, layer-wise LR adjustment derived from layer-specific signals that are extracted locally, without global coordination.

\begin{figure*}[t]
\centering
\includegraphics[width=1.0\linewidth]{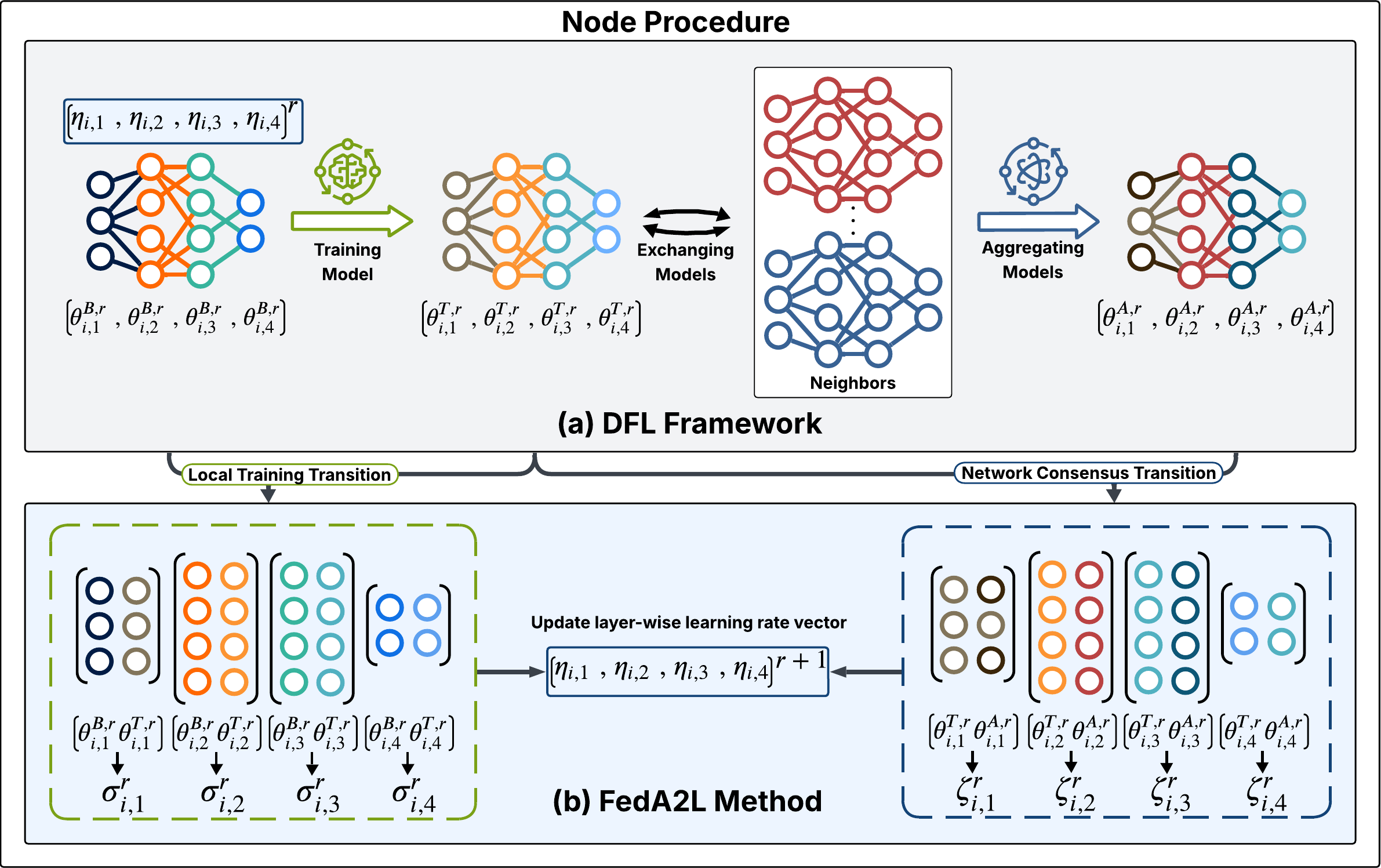} 
\caption{
    (a) Standard DFL framework with three intra-round model states per layer ($\theta_{i,l}^{B,r}$, $\theta_{i,l}^{T,r}$, $\theta_{i,l}^{A,r}$). (b) FedA2L integrated into the DFL framework. During each round, the three states of layer $l$ are used to compute two layer-wise metrics, $\sigma_{i,l}^r$ (Eq.~\ref{eq:sigma_wd}) and $\zeta_{i,l}^r$ (Eq.~\ref{eq:zeta_stability}), which in turn determine the adaptive layer-wise learning rate vector $\eta_i^{r+1} = [\eta_{i,1},\dots,\eta_{i,L}]^{r+1}$ for the next round.
}

\label{fig:fig1}
\end{figure*}

\subsection{Node-level procedure and layer-wise architecture} \label{sec:procedure}

Standard DFL methods apply a single scalar LR $\eta$ uniformly to all model layers, ignoring their distinct optimization needs. In contrast, the proposed approach assigns a layer-wise LR vector $\eta_i^r = [\eta_{i,1}^r, \ldots, \eta_{i,L}^r]$ at each node $i$ and round $r$, where each component $\eta_{i,l}^r$ represents the LR for layer $l$. This vectorized design allows different layers to adopt LRs that better match their roles: foundational layers can use conservative rates to preserve network consensus, while specialized layers can use larger rates to accelerate adaptation to local training. The core challenge is to compute these per-layer LRs adaptively using only information available locally within each DFL round.

To enable such computation, the standard DFL node procedure is formalized in terms of three intra-round model states, as illustrated in Fig.~\ref{fig:fig1}(a). Within each communication round $r$, node $i$ performs three sequential operations: (1) local training on its private dataset $\mathcal{D}_i$ for $E$ epochs, (2) exchange of updated parameters with neighbors $j \in \mathcal{V}_i$, and (3) aggregation of neighbor models to maintain network consensus. This iterative process repeats over $R$ communication rounds. 

FedA2L characterizes the node procedure using three model states within each round. The base state $\theta_{i,l}^{B,r}$ represents the parameters of layer $l$ at node $i$ before local training in round $r$. After local training on $\mathcal{D}_i$ for $E$ epochs, node $i$ reaches the trained state $\theta_{i,l}^{T,r}$, reflecting local update intensity from node-specific data distributions. This local training transition, $\theta_{i,l}^{B,r} \to \theta_{i,l}^{T,r}$, captures how local optimization pushes the layer toward node-specific optima. After model exchange and aggregation with neighbors $j \in \mathcal{V}_i$, node $i$ obtains the aggregated state $\theta_{i,l}^{A,r}$, incorporating neighbor information. This consensus transition, $\theta_{i,l}^{T,r} \to \theta_{i,l}^{A,r}$, captures network consensus constraints at the layer level. The aggregated state then becomes the base state for the next round ($\theta_{i,l}^{B,r+1} \coloneqq \theta_{i,l}^{A,r}$), ensuring continuity across communication rounds. 

\subsection{FedA2L methodology}

FedA2L dynamically computes and assigns layer-wise LRs from real-time model divergence signals extracted from the three-state model of DFL node procedures. The approach operates entirely locally at each node, requiring no centralized coordination or additional communication, thereby preserving the decentralized architecture of DFL. Fig.~\ref{fig:fig1}(b) illustrates the integrated FedA2L method procedure within the standard DFL workflow shown in Fig.~\ref{fig:fig1}(a). 

To provide a formal abstraction of the overall process, FedA2L can be viewed as a per-node mapping from intra-round model states to layer-wise LRs. Specifically, at each node $i$ and round $r$, the method defines the following mapping:

\begin{equation}
M_i^r \bigl(\theta_i^{B,r}, \theta_i^{T,r}, \theta_i^{A,r}\bigr)
\;\rightarrow\;
\eta_i^{r+1} = \bigl[\eta_{i,1}, \eta_{i,2}, \dots, \eta_{i,L}\bigr]^{r+1}
\end{equation}

This mapping is realized through three sequential steps. First, dual layer-wise metrics are extracted from the intra-round state transitions, capturing both local update intensity and network consensus constraints. Next, these metrics are normalized using recent temporal statistics, filtering transient fluctuations to ensure that LR adjustments respond to persistent optimization trends. Finally, the normalized signals are fused into per-layer adaptive LRs through a bounded mechanism. The following subsections detail each step in turn.

\subsubsection{Step 1: Dual metrics for characterizing layer dynamics}

Two complementary metrics are extracted from the three-state model to capture distinct aspects of layer-wise optimization dynamics. Weight divergence quantifies local update intensity during the local training, while aggregation stability quantifies network consensus constraints during the network consensus. These metrics provide layer-specific diagnostic signals encoding the competing forces of local learning and network consensus. FedA2L leverages these signals to compute per-layer LRs balancing both requirements.

\textbf{Local update intensity via weight divergence.} 
To quantify how strongly each layer responds to node-specific data, we examine the parameter change during the local training transition. This transition reveals the degree to which local optimization on node-specific data modifies each layer's parameters. After node $i$ completes $E$ epochs of local training, transitioning from the base state $\theta_{i,l}^{B,r}$ to the trained state $\theta_{i,l}^{T,r}$, the per-layer local update intensity is defined as:

\begin{equation}
    \sigma_{i,l}^{r} = \frac{\| \theta_{i,l}^{T,r} - \theta_{i,l}^{B,r} \|_2}{\| \theta_{i,l}^{B,r} \|_2 + \epsilon}
    \label{eq:sigma_wd}
\end{equation}
where $\|\cdot\|_2$ is the L2-norm and $\epsilon$ is a small constant for numerical stability. Large values of $\sigma_{i,l}^{r}$ reflects substantial parameter modifications during local training, indicating strong adaptation to node-specific data characteristics. Conversely, small values indicate stable layer parameters with minimal changes, suggesting the layer is already well-adapted to local data. This normalized ratio design is scale-invariant, preventing bias from layers with larger absolute parameter magnitudes, and provides a principled measure of relative update magnitude across heterogeneous layer architectures. This per-layer perspective on weight divergence adapts concepts from prior FL work \citep{zhao2018federated} into a diagnostic signal that operates at layer granularity rather than at the model level.

\textbf{Network consensus constraints via aggregation stability.}
To quantify how well each layer aligns with neighbors during the network consensus transition, we examine the magnitude of parameter changes that occur when node $i$ aggregates trained parameters with its neighbors, transitioning from the trained state $\theta_{i,l}^{T,r}$ to the aggregated state $\theta_{i,l}^{A,r}$. The aggregation stability metric measures the proportion of layer parameters that remain stable after aggregation, indicating consensus alignment. Motivated by prior work on consensus mechanisms \citep{zheng2024federated}, the aggregation stability for layer $l$ is defined as:

\begin{equation}
    \zeta_{i,l}^{r} = \frac{1}{|\theta_{i,l}^r|} \sum_{k=1}^{|\theta_{i,l}^r|} \mathbb{I}\left[ \left| \theta_{i,l,k}^{A,r} - \theta_{i,l,k}^{T,r} \right| < \tau \right]
    \label{eq:zeta_stability}
\end{equation}
where $|\theta_{i,l}^r|$ denotes the total number of parameters in layer $l$, $\theta^{T,r}_{i,l,k}$ and $\theta^{A,r}_{i,l,k}$ represent the $k$-th parameter of layer $l$ at node $i$ after local training and after aggregation in round $r$, respectively, $\tau$ is a stability threshold determining parameter stability tolerance, and $\mathbb{I}[\cdot]$ is the indicator function. The metric counts parameters with change magnitude less than $\tau$, normalized by total parameter count. 

A high $\zeta_{i,l}^r$ value (approaching 1) indicates most parameters remained stable during aggregation, reflecting strong consensus alignment and minimal disagreement among neighbors. Conversely, a low value indicates substantial parameter corrections during aggregation, suggesting weak consensus alignment that may require stabilization in subsequent rounds.  

\subsubsection{Step 2: Z-score-based normalization for anomaly detection}

These raw metrics ($\sigma_{i,l}^{r}$ and $\zeta_{i,l}^{r}$) can fluctuate significantly due to stochastic gradient variations and transient network dynamics, particularly during early training when limited historical information is available. To prevent LR adjustments from reacting to noise rather than genuine optimization dynamics, FedA2L employs two mechanisms that stabilize metric estimation: a warm-up initialization period and temporal normalization.

For a warm-up initialization, the base LR is applied uniformly across all layers during the initial $R_{\text{warm}}$ rounds, allowing sufficient metric history to accumulate and enabling early consensus to stabilize. This warm-up period provides a stable foundation for reliable metric estimation, after which FedA2L transitions to temporal normalization to refine LR modulation.

After the warm-up period, FedA2L transitions to temporal normalization to ensure that LR adjustments respond to persistent trends rather than transient fluctuations. For rounds $r > R_{\text{warm}}$, raw metrics are standardized using Z-score normalization computed over a sliding temporal window of the last $\rho$ rounds. This process produces the normalized intensity $\omega^{r}_{i,l}$ (derived from $\sigma_{i,l}$) and $\delta_{i,l}^{r}$ (derived from $\zeta_{i,l}$) signals, defined as:

\begin{equation}
    \omega_{i,l}^{r} = \frac{\sigma_{i,l}^{r} - \frac{1}{\rho}\sum_{k=r-\rho}^{r}\sigma_{i,l}^{k}}
    {\text{STD}(\{\sigma_{i,l}^{k}\}_{k=r-\rho}^{r}) + \epsilon}
    \label{eq:omega_zscore_window}
\end{equation}
\begin{equation}
    \delta_{i,l}^{r} = \frac{\zeta_{i,l}^{r} -  \frac{1}{\rho}\sum_{k=r-\rho}^{r}\zeta_{i,l}^{k}}{\text{STD}(\{\zeta_{i,l}^{k}\}_{k=r-\rho}^{r}) + \epsilon}
    \label{eq:delta_zscore_window}
\end{equation}
where $\text{STD}(\cdot)$ computes the sample standard deviation over the last $\rho$ rounds. These Z-scores quantify deviation from recent historical patterns in standardized units: large positive values indicate abnormal behavior compared to recent history, while values near zero reflect typical and stable behavior. This normalization ensures that LR adjustments respond to meaningful changes in optimization dynamics rather than absolute metric magnitudes, improving robustness across varying network heterogeneity and data distributions. By expressing each metric as a deviation from its own recent history, this step also places $\sigma_{i,l}^r$ and $\zeta_{i,l}^r$ on a comparable scale across layers, which is examined further in the ablation study of Section~\ref{sec:ablation}.

\subsubsection{Step 3: Adaptive layer-wise learning rate} \label{sec:fedA2L_lr}
To reflect local update intensity and network consensus constraints simultaneously, FedA2L synthesizes the normalized signals into a unified fusion score. This fusion process leverages an exponential transformation to assign higher weights to larger deviations while ensuring that all weights remain positive. The fusion score is computed as:

\begin{equation}
    \lambda_{i,l}^{r} = \beta \cdot e^{\omega_{i,l}^{r}} + (1 - \beta) \cdot e^{\delta_{i,l}^{r}}
    \label{eq:fusion}
\end{equation}
where $\beta \in [0,1]$ governs the relative emphasis between local update intensity (first term) and network consensus constraints (second term). This weighted combination produces a positive-valued fusion score that integrates both optimization signals into a single, interpretable measure.

Then, this fusion score is applied to modulate the base LR $\eta^0$ by combining with a global decay factor $\gamma^r = (1 + \xi \cdot r)^{-0.5}$ to produce layer-wise adaptive LRs. The LR for layer $l$ is defined as:

\begin{equation}
    \eta_{i,l}^{r} = \eta^0 \cdot \left(1 + \tanh(\log(\lambda_{i,l}^{r}))\right) \cdot \gamma^r
    \label{eq:final_lr_per_layer}
\end{equation}

where $\xi$ controls the convergence speed across rounds, and the logarithmic transformation stabilizes the range of $\lambda_{i,l}^{r}$ by compressing large variations that could lead to unstable updates. The $\tanh$ function provides smooth, bounded modulation, and the decay factor $\gamma^r$ applies polynomial decay, gradually reducing LRs across rounds while still allowing rapid adaptation in the early stages of training.

Since $\lambda_{i,l}^r>0$, $\log(\lambda_{i,l}^r)\in\mathbb{R}$ and $\tanh(\log(\lambda_{i,l}^r))\in(-1,1)$, which implies $1+\tanh(\log(\lambda_{i,l}^r))\in(0,2)$. Together with $\gamma^r\leq1$, the effective layer-wise LRs satisfy $0<\eta_{i,l}^r<2\eta^0\gamma^r$, so FedA2L cannot arbitrarily amplify the step size. Moreover, the Z-score normalization in Eqs.~\ref{eq:omega_zscore_window} and~\ref{eq:delta_zscore_window} is computed over a sliding window of $\rho$ rounds; persistent divergence shifts the running mean and variance, driving the normalized scores back toward zero and pulling $\lambda_{i,l}^r$ (and thus the LR multiplier) toward $1$ over time.

When the fusion score $\lambda_{i,l}^{r} = 1$ (typical behavior), the modulation factor equals 1, giving $\eta_{i,l}^{r} = \eta^0 \cdot \gamma^r$, which aligns with the global decay schedule and represents a steady-state condition. For $\lambda_{i,l}^{r} > 1$ (high update intensity or weak consensus), the modulation factor exceeds 1, yielding $\eta_{i,l}^{r} > \eta^0 \cdot \gamma^r$ and allowing faster adaptation to node-specific patterns. In contrast, when $\lambda_{i,l}^{r} < 1$ (low update intensity or strong consensus), the LR decreases below the base rate, $\eta_{i,l}^{r} < \eta^0 \cdot \gamma^r$, preserving stable shared representations and network consensus. 

FedA2L interprets large deviations in the normalized metrics $\omega_{i,l}^r$ (derived from $\sigma_{i,l}^r$) and $\delta_{i,l}^r$ (derived from $\zeta_{i,l}^r$) as evidence that a layer is under-adapting to its local objective in non-IID settings. Intuitively, reducing the LR under large deviations may suppress the magnitude of these normalized metrics; however, it also suppresses necessary adaptation in non-IID settings and limits layer-wise specialization. Therefore, when these normalized metrics indicate unusually strong local updates or weak consensus, FedA2L assigns a temporarily higher LR to that layer to accelerate adaptation under non-IID data. 

Importantly, the bounded multiplier $1+\tanh(\log \lambda_{i,l}^r) \in (0,2)$, the global decay factor $\gamma^r$, and the sliding-window Z-score normalization ensure that persistent divergence is re-centered and the effective LR is gradually pulled back toward the global schedule. This fine-grained modulation directly addresses the asymmetric layer-wise dynamics by enabling specialized layers to adapt when local signals are strong, while allowing foundational layers to stabilize when consensus signals are strong. 

\subsection{FedA2L algorithm}
\begin{algorithm}[ht]
\caption{FedA2L method integrated into a DFL framework.}
\label{al:feda2l}
\small

\textbf{Initialize:} For each node $i \in \mathcal{N}$, initialize base model $\theta_i^{B,r}$ at round $r$, local dataset $\mathcal{D}_i$, neighbor list $\mathcal{V}_i$, and history lists $\sigma_i, \zeta_i \gets \emptyset$.

\For{$r = 1$ \KwTo $R$}{
    \For{each node $i \in \mathcal{N}$}{
        $\theta_i^{T,r} \gets \text{LocalTraining}(\theta_i^{B,r}, \mathcal{D}_i, \eta_i^{r})$\;
        Exchange $\theta_i^{T,r}$ with neighbors $\mathcal{V}_i$\;
        $\theta_i^{A,r} \gets \text{Aggregation}(\{\theta_j^{T,r}\}_{j \in \mathcal{V}_i \cup \{i\}})$\;
        $\theta_i^{B,r+1} \gets \theta_i^{A,r}$\\
        \tcp{\textbf{FedA2L: Adaptive LR Computation} }
        \For{each layer $l=1$ \KwTo $L$}{
            $\sigma_{i,l}^{r} \gets \frac{\| \theta_{i,l}^{T,r} - \theta_{i,l}^{B,r} \|_2}{\| \theta_{i,l}^{B,r} \|_2 + \epsilon}$\\
            $\zeta_{i,l}^{r} \gets \frac{1}{|\theta_{i,l}^r|} \sum_{k=1}^{|\theta_{i,l}^r|} \mathbb{I}\left[ \left| \theta_{i,l,k}^{A,r} - \theta_{i,l,k}^{T,r} \right| < \tau \right]$\\
            Update metric histories $\sigma_i, \zeta_i$\; 
            \If{round $r > R_{\text{warm}}$}{
                Compute Z-scores $\omega_{i,l}^{r}$ using Eq.~\ref{eq:omega_zscore_window}\;
                Compute Z-scores $ \delta_{i,l}^{r}$ using Eq.~\ref{eq:delta_zscore_window}\;
                $\lambda_{i,l}^{r} \gets \beta e^{\omega_{i,l}^{r}} + (1 - \beta) e^{\delta_{i,l}^{r}}$\\
                $\eta_{i,l}^{r+1} \gets \eta^0 (1 + \tanh(\log(\lambda_{i,l}^{r}))) \gamma^r$ \\
            }           
        }

    }
    
}
\end{algorithm}

Algorithm~\ref{al:feda2l} presents the complete FedA2L method integrated modularly into a standard DFL training workflow over $R$ communication rounds. At each round $r$, after local training and network consensus, FedA2L executes locally at each node. It computes layer-wise metrics $\sigma_{i,l}^{r}$ and $\zeta_{i,l}^{r}$ from locally available model states ($\theta_i^{B,r}$, $\theta_i^{T,r}$, $\theta_i^{A,r}$) and then produces the adaptive LR vector $\eta_{i,l}^{r+1}$ for the next round. This local-only execution preserves decentralization and does not introduce additional data exposure beyond the standard DFL model-sharing protocol.

FedA2L achieves high computational efficiency, with the per-round and per-node computation having complexity $ \mathcal{O}(|\theta| \cdot |\mathcal{D}_i| \cdot E) $, where $|\theta|$ denotes the model size, \( |\mathcal{D}_i| \) is the dataset size, and $E$ is the number of local steps. In addition, adaptive computation involves per-layer metrics (\( \sigma_{i,l}^r \), \( \zeta_{i,l}^r \)) and historical statistics, with complexity \( \mathcal{O}(|\theta| + L \rho) \), where \( L \) is the number of layers and \( \rho \) is the normalization window size. As the adaptive computation relies solely on model states rather than training data, this overhead scales only with model depth and normalization history, independent of the local dataset size and training workload. Since $|\mathcal{D}_i| \cdot E \gg  L \rho$ in practical DFL configurations, the resulting overhead remains limited compared with the dominant cost of local training. Critically, FedA2L introduces \textit{zero additional communication overhead}, requiring no extra messages, proxy datasets, or centralized coordination, and preserving the decentralized nature and communication efficiency of DFL. 

\section{Convergence analysis}
\label{sec:theory}

We now analyze FedA2L in the DFL setup of Section~\ref{sec:dfl_setup}.
Recall that the network optimizes
\begin{equation}
\min_{\theta_1,\dots,\theta_{|\mathcal{N}|}}
\;\mathcal{F}(\theta_1,\dots,\theta_{|\mathcal{N}|})
\;\triangleq\;
\frac{1}{|\mathcal{N}|}\sum_{i\in\mathcal{N}}\mathcal{L}_i(\theta_i),
\label{eq:network-objective}
\end{equation}
where $\mathcal{L}_i(\theta_i) = \mathbb{E}_{(x,y)\sim\mathcal{D}_i}[\ell(\theta_i;x,y)]$ is the local loss at node $i$.
For the analysis, we also consider the centralized surrogate
\begin{equation}
f(\theta) \triangleq \frac{1}{|\mathcal{N}|}\sum_{i\in\mathcal{N}}\mathcal{L}_i(\theta),
\end{equation}
and track the average model
\begin{equation}
\bar{\theta}^r \triangleq \frac{1}{|\mathcal{N}|}\sum_{i\in\mathcal{N}}\theta_i^{B,r},
\end{equation}
where $\theta_i^{B,r}$ denotes the base state of node $i$ at round $r$ (Section~\ref{sec:procedure}).

Within each round, FedA2L modifies only the local step sizes.
If the model is decomposed into $L$ layers, $\theta_i^{B,r} = [\theta_{i,1}^{B,r};\dots;\theta_{i,L}^{B,r}]$,
the effective LR for layer $l$ at node $i$ in round $r$ is given by Eq.~\ref{eq:final_lr_per_layer}:
\begin{equation}
\eta_{i,l}^r
=
\eta_0\,\gamma_r
\Bigl(1 + \tanh\bigl(\log \lambda_{i,l}^r\bigr)\Bigr),
\qquad
\gamma_r = (1 + \xi r)^{-1/2},
\label{eq:fedA2L-eta}
\end{equation}
where $\lambda_{i,l}^r>0$ is the fusion score computed from the dual metrics
($\sigma_{i,l}^r$ and $\zeta_{i,l}^r$) in Eqs.~\ref{eq:sigma_wd}-\ref{eq:fusion}.
Since $\tanh:\mathbb{R}\to(-1,1)$ and $\lambda_{i,l}^r>0$, we always have
\begin{equation}
0 < \eta_{i,l}^r < 2\eta_0\gamma_r
\quad\text{for all }i,l,r,
\label{eq:eta-bounds}
\end{equation}
i.e., FedA2L keeps all layer-wise LRs strictly positive and uniformly bounded.

\paragraph{Assumptions.}
We adopt standard assumptions from decentralized optimization and DFL:

\textbf{A1 (Smoothness).}
Each $\mathcal{L}_i$ is $\mathbb{L}$-smooth:
$\|\nabla \mathcal{L}_i(x) - \nabla \mathcal{L}_i(y)\| \le \mathbb{L}\|x-y\|$ for all $x,y$.
Hence $f$ is also $\mathbb{L}$-smooth.

\textbf{A2 (Unbiased gradients \& bounded variance).}
Stochastic gradients $g_i(\theta;b)$ (where $b$ denotes a random mini-batch drawn from $\mathcal{D}_i$) satisfy
\[
\mathbb{E}[g_i(\theta;b)] = \nabla \mathcal{L}_i(\theta),
\qquad
\mathbb{E}\bigl[\|g_i(\theta;b)- \nabla \mathcal{L}_i(\theta)\|^2\bigr] \le \varrho^2.
\]

\textbf{A3 (Mixing matrix \& spectral gap).}
Each mixing matrix $W^r = [w_{ij}^r]_{i,j}$ used in the aggregation step
(Algorithm~\ref{al:feda2l}, line~6) is symmetric and doubly stochastic,
$W^r\mathbf{1}=\mathbf{1}$ and $(W^r)^\top\mathbf{1}=\mathbf{1}$,
and there exists a spectral gap $\phi\in(0,1]$ such that
\[
\bigl\|W^r - \tfrac{1}{|\mathcal{N}|}\mathbf{1}\mathbf{1}^\top\bigr\|_2 \le 1-\phi
\quad\text{for all }r.
\]

\textbf{A4 (Data heterogeneity).}
There exists $B\ge 0$ such that for all $\theta$,
\[
\frac{1}{|\mathcal{N}|}\sum_{i\in\mathcal{N}}
\bigl\|\nabla \mathcal{L}_i(\theta) - \nabla f(\theta)\bigr\|^2 \le B^2.
\]

\textbf{A5 (Bounded and slowly varying multipliers).}
Define layer-wise multipliers $m_{i,l}^r$ via
$\eta_{i,l}^r = m_{i,l}^r (\eta_0\gamma_r)$, where $\eta_0 = c/\sqrt{R}$.
There exist constants $0 < m_{\min} \le m_{\max} < \infty$ and $\Delta_m<\infty$ such that
\[
m_{\min} \le m_{i,l}^r \le m_{\max},
\qquad
|m_{i,l}^{r+1} - m_{i,l}^r| \le \Delta_m,
\quad \forall\,i,l,r.
\]
In FedA2L, \(m_{i,l}^r = 1 + \tanh(\log \lambda_{i,l}^r)\), so \(0 < m_{i,l}^r < 2\) by design, and the Z-score based update over a finite window (Eqs.~\ref{eq:omega_zscore_window}-\ref{eq:delta_zscore_window}) bounds metric volatility, which, when coupled with the bounded \(\tanh\) transformation, ensures that \(|m_{i,l}^{r+1} - m_{i,l}^r| \le \Delta m\) for some finite \(\Delta m\).

We also measure consensus by the mean-squared disagreement
\begin{equation}
D^r \triangleq \frac{1}{|\mathcal{N}|}\sum_{i\in\mathcal{N}}\sum_{l=1}^L
\bigl\|\theta_{i,l}^{B,r} - \bar{\theta}_l^r\bigr\|^2,
\qquad
\bar{\theta}_l^r = \frac{1}{|\mathcal{N}|}\sum_{i\in\mathcal{N}}\theta_{i,l}^{B,r}.
\end{equation}

\subsection{Non-convex convergence}

We first consider the general non-convex setting. Let $\eta_{\min}^r \le \eta_{i,l}^r \le \eta_{\max}^r$ denote lower/upper bounds on the effective step sizes in round $r$, induced by \eqref{eq:eta-bounds} and A5.

\begin{theorem}[Non-convex FedA2L convergence]
\label{thm:fedA2L-nonconvex}
Assume \textbf{A1}-\textbf{A5} and one local SGD epoch per round at each node (Algorithm~\ref{al:feda2l} with $E=1$). Suppose the effective step sizes satisfy \(0 < \eta_{\min}^r \le \eta_{i,l}^r \le \eta_{\max}^r\) with \(\eta_{\max}^r L\) sufficiently small for all \(r\), which is the standard small-step condition ensuring that higher-order perturbation terms remain controlled under Lipschitz smoothness. Then, for any $R \ge 1$,

\begin{equation}
\begin{aligned}
\frac{\sum_{r=0}^{R-1} \eta_{\min}^r \,
\mathbb{E}\bigl[\|\nabla f(\bar{\theta}^r)\|^2\bigr]}
{\sum_{r=0}^{R-1} \eta_{\min}^r}
&\le
\frac{2\bigl(f(\bar{\theta}^0) - f_\star\bigr)}
{\sum_{r=0}^{R-1}\eta_{\min}^r}
+ C_1 L\,\overline{\eta}_{\max}\,\frac{\varrho^2}{|\mathcal{N}|} \\
&\quad
+ C_2 \frac{L^2\,\overline{\eta}_{\max}^2 B^2}{\phi^2}
+ C_3 \frac{L\,\overline{\eta}_{\max}^2 \Delta_m^2}{\phi^2},
\end{aligned}
\label{eq:fedA2L-nonconvex-bound}
\end{equation}
where $f_\star = \inf_\theta f(\theta)$, $\overline{\eta}_{\max} = \max_{0\le r<R}\eta_{\max}^r$, and $C_1,C_2,C_3>0$ are constants independent of $R$.
\end{theorem}

\noindent\emph{Sketch of proof.}
Using $\mathbb{L}$-smoothness, one shows that the average model satisfies
\[
\begin{aligned}
\mathbb{E}[f(\bar{\theta}^{r+1})]
&\le \mathbb{E}[f(\bar{\theta}^r)]
- \tfrac{\eta_{\min}^r}{2}\,\mathbb{E}\|\nabla f(\bar{\theta}^r)\|^2 \\
&\quad + O\!\left((\eta_{\max}^r)^2
\bigl(\varrho^2 + B^2 + \mathbb{E}D^r\bigr)\right).
\end{aligned}
\]
A separate recursion for the disagreement $D^r$ follows from the spectral gap $\phi$ of $W^r$: consensus contracts by a factor $(1-\phi)$ each round, but is driven by stochastic gradient noise $\varrho^2$, heterogeneity $B^2$, and the per-round drift $\Delta_m$ of the multipliers, which remains bounded due to the multiplier construction in A5 and the bounded tanh transformation in Eq.~\ref{eq:final_lr_per_layer}. Summing the descent inequality over $r = 0,\dots,R-1$, bounding $\mathbb{E}D^r$ by the steady state of this consensus recursion, and dividing by $\sum_{r=0}^{R-1}\eta_{\min}^r$ yields~\eqref{eq:fedA2L-nonconvex-bound}.

\paragraph{Diminishing step sizes.}
For the theoretical analysis, we set $\gamma_r = 1$ and use the budget-dependent base LR
$\eta_0 = c/\sqrt{R}$ from A5. Combined with bounded multipliers
$m_{i,l}^r\in[m_{\min},m_{\max}]$ (A5), this gives round-independent effective step sizes $\eta_{i,l}^r = m_{i,l}^r\,\eta_0$, so
\[
\begin{aligned}
  \overline{\eta}_{\max} &= m_{\max}\,\frac{c}{\sqrt{R}} = O\!\left(\frac{1}{\sqrt{R}}\right), \\
  \sum_{r=0}^{R-1}\eta_{\min}^r &\ge R\cdot m_{\min}\,\frac{c}{\sqrt{R}} = \Omega(\sqrt{R}).
\end{aligned}
\]
The three error terms in~\eqref{eq:fedA2L-nonconvex-bound} then satisfy
$C_1 L\,\overline{\eta}_{\max} = O(1/\sqrt{R})$ and
$C_2,C_3\propto\overline{\eta}_{\max}^2 = O(1/R)$, so all terms vanish as $R\to\infty$.
Substituting into Theorem~\ref{thm:fedA2L-nonconvex} yields
\[
  \frac{\sum_{r=0}^{R-1}\eta_{\min}^r\,\mathbb{E}\|\nabla f(\bar{\theta}^r)\|^2}
       {\sum_{r=0}^{R-1}\eta_{\min}^r}
  = O\!\left(\frac{1}{\sqrt{R}}\right),
\]
matching the standard $O(1/\sqrt{R})$ non-convex rate of decentralized SGD. FedA2L does
not worsen the convergence order; the bounded multipliers $m_{i,l}^r$ only modify
the constants $C_1,C_2,C_3$. In practice we employ the diminishing schedule
$\gamma_r=(1+\xi r)^{-1/2}$ to improve empirical stability.

\subsection{Strongly convex case}
If $f$ is additionally $\mu$-strongly convex and the effective step sizes are kept constant over rounds, $\eta_{i,l}^r \in [\eta_{\min},\eta_{\max}]$ with $\eta_{\max}$ sufficiently small, the same arguments with a Lyapunov function $V^r = \mathbb{E}[f(\bar{\theta}^r)-f_\star] + \gamma\,\mathbb{E}[D^r]$ show that FedA2L preserves the linear convergence of decentralized SGD: for suitable $\gamma>0$,
\begin{equation}
\begin{aligned}
&\mathbb{E}\bigl[f(\bar{\theta}^R) - f_\star\bigr]
\le
\rho_0^R\bigl(f(\bar{\theta}^0) - f_\star\bigr) \\
&\quad+ O\!\Biggl(
\frac{L\,\overline{\eta}_{\max}\varrho^2}{|\mathcal{N}|}
+ \frac{L^2\,\overline{\eta}_{\max}^2 B^2}{\phi^2}
+ \frac{L\,\overline{\eta}_{\max}^2 \Delta_m^2}{\phi^2}
\Biggr),
\end{aligned}
\label{eq:str-convex}
\end{equation}
with contraction factor $\rho_0 \in (0,1)$ determined jointly by $\mu$ and the spectral gap $\phi$. Thus, in the constant step-size regime, FedA2L does not degrade the qualitative linear rate of the underlying DFL algorithm; it only changes the constants through bounded, layer-wise multipliers.

This strongly convex result is provided for completeness in the constant step-size regime. Since our experiments adopt the diminishing schedule in Eq.~\ref{eq:fedA2L-eta}, the main theoretical implication for the practical setup is the preservation of the standard non-convex convergence rate established above.

\section{Experiments} \label{sec:exp}

This section presents the evaluation setup and performance analysis of FedA2L across image classification and time-series forecasting (TSF) tasks, covering multiple benchmark datasets, model architectures, DFL algorithms, and network configurations. The evaluation considers five complementary dimensions. Convergence speed is measured by communication rounds required to reach a target accuracy. Computational efficiency is measured by total time to reach the same target. Model quality is assessed by best test accuracy on classification tasks and MSE on regression tasks, capturing whether convergence gains carry through to model performance. Robustness is examined under varying data heterogeneity, network scales, and sparse topologies. Finally, ablation and sensitivity studies isolate the contribution of each methodological component and examine how key design choices affect convergence behavior. The corresponding analyses are presented in Sections \ref{sec:fastconv}-\ref{sec:ablation}.

\subsection{Experimental setup}
\subsubsection{Datasets}

Experimental evaluation was conducted on five benchmark datasets spanning image classification and TSF. For image classification, three standard benchmarks are used: CIFAR-10, CIFAR-100, and TinyImageNet. CIFAR-10 and CIFAR-100 \citep{krizhevsky2009learning} contain 60,000 32$\times$32 color images across 10 and 100 classes, respectively, enabling assessment of both coarse and fine-grained classification tasks. TinyImageNet \citep{le2015tiny} comprises 100,000 64$\times$64 color images across 200 classes, providing a more challenging and realistic benchmark for evaluating scalability on larger-scale vision tasks.

To emulate realistic heterogeneous data distributions (non-IID data distribution), a Dirichlet distribution with parameter $\alpha$ was used to partition datasets across nodes. Each node $i$ receives a subset of samples, where $\alpha$ controls heterogeneity level. Lower $\alpha$ values correspond to severe non-IID conditions where each node receives highly skewed class distributions, while higher values yield more balanced distributions approaching IID. Default experiments used $\alpha = 0.1$, with robustness analysis varying $\alpha \in \{0.01, 0.5\}$.

For TSF, two real-world benchmarks are used: ExchangeRate \citep{lai2018modeling} and BeijingAirQuality \citep{zhang2017cautionary}. ExchangeRate contains 7,588 daily exchange rate records across 8 currencies, where each currency corresponds to a node. BeijingAirQuality comprises hourly air quality measurements from 12 monitoring stations across Beijing, with 11 variates per node. Both use an input horizon of 96 and a prediction horizon of 96 with naturally defined node partitions (i.e., no Dirichlet-based partitioning). These datasets reflect real-world distributed time-series data with temporal dependencies across nodes. 

\subsubsection{Models}
Model architectures cover varying complexity levels. For image classification, a 4-layer convolutional neural network (CNN) for lightweight evaluation, ResNet-18 for moderate-scale tasks, and ResNet-34 for deeper network evaluation. For TSF, we use Timer~\citep{10.5555/3692070.3693383}, an 18-layer transformer-based forecasting model. This diversity of models enables a comprehensive evaluation of FedA2L's effectiveness across classification and regression settings with architectures of different depth and complexity.

\subsubsection{DFL configuration}
All experiments were conducted with $|\mathcal{N}| = 10$ nodes arranged in a fully connected P2P topology, using synchronous communication unless otherwise specified. Each node performs local training for $E = 1$ epoch per communication round, with initial LR $\eta^0 = 0.01$. Classification experiments train for $R=500$ communication rounds, and convergence is evaluated by measuring the rounds required to reach predefined target test accuracy. The target accuracies are selected as stable and practically meaningful convergence levels that the majority of baseline methods can reach and sustain, reflecting genuine learning progress in the distributed setting. TSF experiments train for $R=100$ communication rounds, and convergence is evaluated by the rounds required to reach a target MSE threshold. All experiments are conducted on a server equipped with an NVIDIA GeForce RTX 4090 GPU and 126 GB of RAM, running PyTorch 2.5.0 under Python 3.12. All results represent mean values across 5 independent runs with different random seeds to ensure statistical robustness.

\subsubsection{Baseline methods}
FedAvg \citep{mcmahan2017communication} (vanilla averaging baseline), FedProx \citep{li2020federated} (proximal term for heterogeneity), FedYogi \citep{reddi2021adaptive} (adaptive second-moment optimization), FedNTD \citep{lee2021fedntd} (knowledge distillation approach), FedAWA \citep{shi2025fedawa} (adaptive aggregation method), DFedSAM~\citep{DFedSAM} (sharpness-aware minimization for flat minima under non-IID conditions), and DFedHPO~\citep{DFedHPO} (decentralized hyperparameter optimization). DFedHPO conducts a one-time decentralized search to provide a single optimal fixed LR configuration before training begins, while FedA2L continuously adapts layer-wise LRs during training. As a result, the fixed LR from DFedHPO serves as an initialization, and subsequent optimization is governed by FedA2L. Accordingly, DFedHPO is evaluated as a standalone reference baseline, using its searched LR as a fixed configuration without additional tuning procedures or schedulers. This enables a direct comparison between static pre-training hyperparameter search and dynamic layer-wise adaptation during training. These baselines represent classical, regularized, adaptive, network-aware, geometric, and hyperparameter optimization strategies in DFL, providing a solid foundation for comparison with FedA2L, a layer-wise adaptive LR method. For each base algorithm, three distinct LR strategies are evaluated, with the exception of DFedHPO, which is assessed only under its searched LR configuration as described above: 

\begin{itemize}
    \item \textbf{Vanilla (Unchanged LR):} Single constant LR maintained throughout training, serving as the reference for comparison.
    \item \textbf{Scheduler-based:} Global LR adjusted per round according to established schedulers: StepLR (exponential decay), CosineAnnealingWarmRestarts (CAWR) with periodic warm restarts \citep{loshchilov2017sgdrstochasticgradientdescent}, OneCycleLR (OCLR) \citep{smith2018superconvergencefasttrainingneural} with triangular scheduling, and Hyperbolic scheduler \citep{kim2025hyperboliclrepochinsensitivelearning} providing efficient decay. 
    \item \textbf{FedA2L (Proposed):} Layer-wise adaptive LRs dynamically computed per round. 
\end{itemize}

These three strategy categories encompass the spectrum from static (vanilla) through globally adaptive (schedulers) to locally adaptive layer-wise (FedA2L), providing a comprehensive evaluation framework. This evaluation structure enables fair assessment: unchanged and scheduler baselines provide performance bounds for global (non-adaptive) strategies, while FedA2L demonstrates the advantage of per-layer adaptation. 

\subsection{How fast does FedA2L converge?} \label{sec:fastconv}
\begin{table*}[!h]
\centering
\caption{Communication rounds to target accuracy in DFL (FedA2L vs. baselines). Results use a sliding average (window = 10). Bold: fastest; underline: second-fastest; “\_”: no convergence. `$1^{st}$ count' = number of fastest cases. DFedHPO is reported only under its searched LR, as it does not incorporate additional LR scheduling.} 
\label{tab:main_results}
\resizebox{0.98\linewidth}{!}{
\begin{tabular}{cc|cc|cc|cc|cc}
\toprule
\multicolumn{2}{c|}{Model} & \multicolumn{2}{c|}{CNN} & \multicolumn{2}{c|}{ResNet-18} & \multicolumn{2}{c|}{ResNet-34}& \multirow{2}{*}{\makecell{$1^{st}$ \\ Count}} \\
\cmidrule(lr){1-8}
\multicolumn{2}{c|}{Dataset} 

& CIFAR-10 & CIFAR-100 
& CIFAR-100 & TINY 
& CIFAR-100 & TINY \\
\cmidrule(lr){1-2} \cmidrule(lr){3-8}
 Algorithm & Method
& 64\% & 29\% 
& 45\% & 35\% 
& 41\% & 30\% & \\
\midrule

\multirow{6}{*}{FedAvg} 
& Vanilla      & 374$\pm$38.93  & 29$\pm$22.07  & 491$\pm$53.13  & \_ & 406$\pm$50.26  & \_ & 0 \\ 
& StepLR       & 190$\pm$23.25  & 27$\pm$0.4  & 309$\pm$33.86  & \underline{314$\pm$44.85 } & 308$\pm$8.52  & 214$\pm$29.17  & 0 \\ 
& CAWR         & 191$\pm$40.63  & \textbf{23$\pm$0.64 } & \underline{197$\pm$46.2 } & \_ & 199$\pm$23.84  & \underline{148$\pm$38.81 } & 1\\ 
& OCLR         & \underline{165$\pm$17.81 } & 58$\pm$1.09  & 266$\pm$11.23  & 450$\pm$9.29  & \underline{142$\pm$9.94 } & 225$\pm$12.53  & 0 \\
& Hyperbolic   & 221$\pm$17.82  & 27$\pm$36.8  & 432$\pm$37.5  & \_ & \_ & 256$\pm$32.93  & 0\\ 
& FedA2L       & \textbf{136$\pm$8.21 } & \textbf{23$\pm$0.49 } & \textbf{177$\pm$6.97 } & \textbf{187$\pm$15.83 } & \textbf{120$\pm$12.1 } & \textbf{71$\pm$1.25 } & \textbf{6} \\ 
& \textit{Improv.} & \textit{17.57\%} & \textit{0.0\%} & \textit{10.15\%} & \textit{40.44\%} & \textit{15.49\%} & \textit{52.03\%} & \\
\midrule

\multirow{6}{*}{FedProx} 
& Vanilla      & 285$\pm$51.04  & 25$\pm$0.8  & \_ & \_ & 389$\pm$75.5  & \_ & 0\\ 
& StepLR       & \underline{144$\pm$14.6 } & 25$\pm$0.8  & \_ & \_ & 406$\pm$66.09  & \_ & 0\\ 
& CAWR         & 148$\pm$64.78  & \textbf{23$\pm$0.63 } & 297$\pm$23.63  & \_ & 244$\pm$41.62  & 194$\pm$22.35  & 1\\  
& OCLR         & 149$\pm$29.97  & 48$\pm$0.8  & \underline{247$\pm$11.38 } & \underline{447$\pm$4.72 } & \underline{132$\pm$10 } & \underline{182$\pm$11.92 } & 0 \\ 
& Hyperbolic   & 213$\pm$27.6  & 25$\pm$0.75  & \_ & \_ & 266$\pm$38.96  & 251$\pm$52.89  & 0\\ 
& FedA2L       & \textbf{127$\pm$24.89 } & \textbf{23$\pm$0.49 } & \textbf{175$\pm$15.14 } & \textbf{185$\pm$7.88 } & \textbf{115$\pm$3.24 } & \textbf{82$\pm$0.82 } & \textbf{6}  \\ 
& \textit{Improv.} & \textit{11.81\%} & \textit{0.0\%} & \textit{29.15\%} & \textit{58.61\%} & \textit{12.88\%} & \textit{54.95\%} & \\
\midrule

\multirow{6}{*}{FedYogi}
& Vanilla      & 311$\pm$34.98  & 27$\pm$1.79  & \_ & \_ & 367$\pm$51.4  & 308$\pm$5.63  & 0\\ 
& StepLR       & 163$\pm$29.92  & 27$\pm$1.79  & \_ & \_ & 306$\pm$51.39  & 207$\pm$27.81  & 0 \\ 
& CAWR         & 189$\pm$35.88  & \textbf{23$\pm$0.63 } & \textbf{198$\pm$48.88 } & \underline{248$\pm$35.39 } & 198$\pm$67.97  & 195$\pm$21.82  & 2 \\ 
& OCLR         & \underline{161$\pm$22.01 } & 49$\pm$0.98  & 273$\pm$8.45  & 454$\pm$11.23  & \underline{152$\pm$17.69 } & \underline{182$\pm$12.03 } & 0  \\  
& Hyperbolic   & 214$\pm$31.26  & 27$\pm$1.09  & 361$\pm$27.6  & \_ & 324$\pm$42.15  & 251$\pm$72.24  & 0 \\ 
& FedA2L       & \textbf{141$\pm$25.82 } & \underline{25$\pm$0.49 } & \underline{210$\pm$8.7 } & \textbf{240$\pm$6.98 } & \textbf{134$\pm$7.678 } & \textbf{74$\pm$1.67 } & \textbf{4} \\
& \textit{Improv.} & \textit{12.42\%} & \textit{-8.69\%} & \textit{-6.06\%} & \textit{3.22\%} & \textit{11.84\%} & \textit{59.34\%} & \\
\midrule

\multirow{6}{*}{FedNTD} 
& Vanilla      & 225$\pm$68.65  & 24$\pm$0.89  & 242$\pm$29.17  & \_ & 239$\pm$45.26  & \_ & 0\\ 
& StepLR       & \textbf{141$\pm$27.91 } & 24$\pm$0.89  & \underline{133$\pm$40.57 } & \_ & 209$\pm$51  & \_ & 1\\ 
& CAWR         & 190$\pm$72.52  & \textbf{21$\pm$0.4 } & 143$\pm$3.06  & \_ & 136$\pm$20.44  & 149$\pm$1.5  & 1\\ 
& OCLR         & \underline{150$\pm$17.74 } & 47$\pm$1.02  & 161$\pm$4.21  & \_ & \underline{99$\pm$1.94 } & \underline{124$\pm$3.29 } & 0  \\  
& Hyperbolic   & 218$\pm$24.68  & 24$\pm$1.2  & 175$\pm$16.02  & \_ & \_ & \_ & 0 \\ 
& FedA2L       & 167$\pm$17.29  & \underline{23$\pm$0.49 } & \textbf{122$\pm$7.81 } & \textbf{175$\pm$17.15 } & \textbf{88$\pm$1.02 } & \textbf{72$\pm$1.69 } & \textbf{4} \\ 
& \textit{Improv.} & \textit{-18.44\%} & \textit{-9.5\%} & \textit{8.3\%} & \textit{100\%} & \textit{11.11\%} & \textit{41.94\%} & \\
\midrule

\multirow{6}{*}{FedAWA} 
& Vanilla      & 387$\pm$51.94  & 30$\pm$29.67  & \_ & \_ & 264$\pm$75  & 356$\pm$74.96  & 0 \\ 
& StepLR       & 195$\pm$14.03  & 27$\pm$53.83  & 409$\pm$49.50  & \_ & 202$\pm$41.41 & 201$\pm$15.92  & 0 \\ 
& CAWR         & 243$\pm$77.98  & \textbf{23$\pm$0.8 } & 247$\pm$68.22  & \underline{248$\pm$41.42 } & 197$\pm$24.31  & \underline{146$\pm$1.25 } & 1 \\ 
& OCLR         & \underline{174$\pm$18.11 } & 48$\pm$1.02  & \underline{244$\pm$17.68 } & 440$\pm$29.56  & \underline{139$\pm$3.76 } & 209$\pm$0.82  & 0 \\  
& Hyperbolic   & 218$\pm$13.95  & 26$\pm$1.74  & 372$\pm$86.86  & \_ & 286$\pm$70.81  & 495$\pm$113.5  & 0\\ 
& FedA2L       & \textbf{152$\pm$14.66 } & \textbf{23$\pm$0.8 } & \textbf{177$\pm$7.49 } & \textbf{195$\pm$8.24 } & \textbf{112$\pm$3.26 } & \textbf{72$\pm$0.94 } & \textbf{6} \\ 
& \textit{Improv.} & \textit{12.64\%} & \textit{0.0\%} & \textit{27.45\%} & \textit{21.37\%} & \textit{19.42\%} & \textit{50.68\%} & \\
\midrule

\multirow{6}{*}{DFedSAM} 
& Vanilla    &346$\pm$55.87   &36$\pm$5.01   &167$\pm$17.59  &254$\pm$17.59  &117$\pm$10.21 &72$\pm$2.16  &0  \\ 
& StepLR     &\underline{160$\pm$24.07 }  &36$\pm$60.81  &142$\pm$22.01  &207$\pm$48.56  &101$\pm$7.76 &73$\pm$4.24  &0  \\ 
& CAWR       &197$\pm$24.09   &\textbf{27$\pm$0.47 }  &97$\pm$22.23   &\underline{84$\pm$1.25 } &92$\pm$2.62 &86$\pm$1.25  &1  \\ 
& OCLR       &170$\pm$29.58   &53$\pm$0.82   &\underline{95$\pm$1.7 }    &94$\pm$2.45  &\underline{85$\pm$1.89} &73$\pm$0.47  &0  \\  
& Hyperbolic &196$\pm$19.13   &\underline{31$\pm$0.47}   &161$\pm$0.47   &175$\pm$55.5  &95$\pm$27.82 &175$\pm$55.75 &0  \\ 
& FedA2L     &\textbf{114$\pm$2.36 }  &\textbf{27$\pm$0.49} &\textbf{88$\pm$1.25 } &\textbf{81$\pm$2.85 } &\textbf{80$\pm$2.62 } &\textbf{70$\pm$0.82}  &\textbf{6}  \\ 
& \textit{Improv.}    &\textit{28.75\%}  &\textit{0.0\%}  &\textit{7.37\%}   &\textit{3.57\%}  &\textit{5.88\%}   &\textit{2.77\%}  &  \\ 
\midrule
{DFedHPO}
& Vanilla    &170$\pm$10.27  &60$\pm$0.82  &\_           &\_      &   243$\pm$10.34   &141$\pm$45.9 &  \\

\bottomrule
\end{tabular}
}
\end{table*}
\subsubsection{Comparison across models and datasets}

FedA2L demonstrates consistent acceleration across diverse models and datasets, with the performance gain increasing as model complexity and dataset diversity increase. Across the 36 scheduler-strategy settings in Table~\ref{tab:main_results}, FedA2L achieves the fastest convergence in 32 cases and the second-fastest in 3, confirming robust and generalizable performance. FedA2L also converges faster than DFedHPO across all evaluated 
configurations, confirming the advantage of dynamic layer-wise 
adaptation over static hyperparameter search. The \textit{Improv.} rows quantify the percentage reduction in communication rounds relative to the best-performing baseline per setting, providing a direct measure of practical efficiency gain. 

Across ResNet-18 and ResNet-34 architectures, FedA2L consistently reduces convergence rounds across all seven algorithms, with the strongest gains on TinyImageNet ranging from 2.77\% under DFedSAM to 59.34\% under FedYogi. Against the best scheduler, FedA2L achieves further reductions on all ResNet settings. Additionally, FedA2L exhibits notably low variance across independent runs: on TinyImageNet with ResNet-34 under FedAvg, FedA2L converges in $71{\pm}1.25$ rounds compared to $148{\pm}38.81$ for CAWR, reflecting stable and reproducible behavior that is important for deployment in resource-constrained edge environments.

On CNN architectures, FedA2L achieves positive gains on CIFAR-10 across most algorithms, reaching up to 28.75\% improvement under DFedSAM, while ranking second in the remaining settings where the base algorithm already partially captures layer-level variance. In all such cases, FedA2L still outperforms the vanilla baseline, confirming that the layer-wise mechanism does not destabilize training. The reasons for these interactions are discussed in the following subsection.

\subsubsection{Comparison across DFL algorithms} \label{sec:algo_comparison}

A key property of FedA2L is its algorithmic orthogonality, enabling seamless integration with existing DFL protocols without modifying the core aggregation logic. This is particularly important for practitioners deploying mature, established DFL systems. Table~\ref{tab:main_results} shows consistent acceleration across six DFL algorithms, confirming that this property holds across fundamentally different DFL design philosophies under diverse update mechanisms. Under FedProx, which applies proximal regularization to reduce client drift, FedA2L still achieves 2.24$\times$ speedup on CNN CIFAR-10 (127 vs. 285 rounds), confirming that layer-wise rate adaptation targets an orthogonal optimization dimension to loss-based regularization. Under FedAWA, which reweighs aggregation contributions based on model similarity, FedA2L reaches 72 rounds versus 356 rounds on ResNet-34 TinyImageNet (4.94$\times$ speedup), demonstrating that adaptive aggregation and adaptive layer-wise LRs are complementary rather than competing mechanisms.

The only exceptions arise in two CNN configurations under FedNTD and two CIFAR-100 settings under FedYogi. In these cases, the base algorithm already partially captures layer-level variance through knowledge distillation in FedNTD and per-parameter moment estimation in FedYogi, reducing the available margin for further layer-wise modulation. This limits, but does not eliminate, the benefit of additional layer-wise adaptation. Nonetheless, FedA2L still outperforms the vanilla baseline in all four cases, confirming that the layer-wise mechanism remains compatible with the base protocol even when divergence signals are partially attenuated.

\subsubsection{Comparison across learning rate strategies}
Global LR schedulers address temporal heterogeneity by adjusting rates over training rounds. However, they cannot address spatial heterogeneity, which refers to the differing optimization needs across model layers. FedA2L consistently outperforms these scheduler-based baselines (Table~\ref{tab:main_results}) by leveraging local layer signals to detect each layer's specific optimization needs during each communication round. This distinction becomes critical in deep heterogeneous models. With ResNet-18 on CIFAR-100 (45\% target accuracy), FedA2L converges in 177 rounds compared to 491 rounds for the vanilla baseline under FedAvg ($2.77\times$ speedup) and 197 rounds for the best scheduler, CAWR (10.15\% further improvement). On TinyImageNet with ResNet-18, the advantage becomes more pronounced. Under FedNTD, all schedulers fail to converge, whereas FedA2L reaches the target in 175 rounds, demonstrating that global scheduling is insufficient under the combined effects of deep layer heterogeneity and severe non-IID conditions.

\subsubsection{Comparison across model architectures}

The effectiveness of FedA2L scales consistently with architectural depth, following a pattern visible across all algorithm rows in Table~\ref{tab:main_results}. On ResNet-34, the deepest architecture evaluated, FedA2L delivers the strongest gains across all algorithms. On TinyImageNet, improvements range from 2.77\% under DFedSAM to 59.34\% under FedYogi, while the vanilla baseline fails to converge in several settings where FedA2L successfully reaches the target accuracy. This pattern confirms that deep feature hierarchies combined with large class diversity produce strong layer-wise divergence signals, precisely the condition under which FedA2L's per-layer rate control is most effective under non-IID settings. On ResNet-18, the reduced depth and lower class diversity yield consistent but more modest improvements across all base algorithms on both CIFAR-100 and TinyImageNet. On CNN, FedA2L acts primarily as a training stabilizer, achieving a $2.75\times$ speedup over vanilla under FedAvg on CIFAR-10 while narrowing scheduler gaps rather than delivering the strongest speedups in every configuration. These results indicate that FedA2L's benefits increase with feature hierarchy depth and representation complexity, rather than with network depth alone.

\begin{figure*}[t]
\centering

\begin{minipage}{\textwidth}
\centering
\captionof{table}{Comparing time to reach 45\% target accuracy (total time in minutes) on CIFAR-100 using ResNet-18 (Dirichlet $\alpha=0.1$).}
\label{tab:con-time}

\vspace{4pt}

\resizebox{\textwidth}{!}{%
\begin{tabular}{c|cc|cc|cc|cc|cc|cc}
\toprule
Algorithm
 & \multicolumn{2}{c|}{FedAvg} & \multicolumn{2}{c|}{FedProx} & \multicolumn{2}{c|}{FedYogi} & \multicolumn{2}{c|}{FedNTD} & \multicolumn{2}{c|}{FedAWA} & \multicolumn{2}{c}{DFedSAM} \\
\cmidrule(lr){1-1}\cmidrule(lr){2-3} \cmidrule(lr){4-5} \cmidrule(lr){6-7} \cmidrule(lr){8-9} \cmidrule(lr){10-11} \cmidrule(lr){12-13}
Method
 & \begin{tabular}[c]{@{}c@{}c@{}}Time per\\round\\ (s)\end{tabular} 
 & \begin{tabular}[c]{@{}c@{}c@{}}Total\\Time\\(min)\end{tabular}

 & \begin{tabular}[c]{@{}c@{}c@{}}Time per\\round\\ (s)\end{tabular} 
 & \begin{tabular}[c]{@{}c@{}c@{}}Total\\Time\\(min)\end{tabular}

 & \begin{tabular}[c]{@{}c@{}c@{}}Time per\\round\\ (s)\end{tabular} 
 & \begin{tabular}[c]{@{}c@{}c@{}}Total\\Time\\(min)\end{tabular}

 & \begin{tabular}[c]{@{}c@{}c@{}}Time per\\round\\ (s)\end{tabular} 
 & \begin{tabular}[c]{@{}c@{}c@{}}Total\\Time\\(min)\end{tabular}

 & \begin{tabular}[c]{@{}c@{}c@{}}Time per\\round\\ (s)\end{tabular} 
 & \begin{tabular}[c]{@{}c@{}c@{}}Total\\Time\\(min)\end{tabular}

 & \begin{tabular}[c]{@{}c@{}c@{}}Time per\\round\\ (s)\end{tabular} 
 & \begin{tabular}[c]{@{}c@{}c@{}}Total\\Time\\(min)\end{tabular} \\
\midrule

Vanilla 
& 28.43 & 232.63 
& 30.02 & \_ 
& 28.62 & \_ 
& 32.69 & 131.85 
& 28.18 & \_
& 46.61 & 129.73 \\

StepLR 
& 28.23 & 145.38 
& 30.02 & \_ 
& 28.60 & \_ 
& 32.46 & \underline{71.95}
& 28.38 & 193.46 
& 46.23 & 67.68 \\

CAWR 
& 28.15 & \underline{92.41} 
& 29.68 & 146.92 
& 28.82 & \textbf{95.09}
& 32.82 & 78.22 
& 28.48 & 117.23 
& 46.47 & 46.59 \\

OCLR 
& 28.15 & 124.78 
& 29.78 & \underline{122.60}
& 28.62 & 130.22 
& 32.35 & 86.81 
& 28.15 & \underline{114.48}
& 46.51 & \underline{45.31} \\

Hyperbolic 
& 28.15 & 202.71 
& 30.29 & \_ 
& 28.70 & 172.65 
& 32.94 & 96.08 
& 28.48 & 176.59 
& 46.74 & 77.00 \\

FedA2L 
& 29.28 & \textbf{86.38}
& 31.47 & \textbf{91.79} 
& 30.73 & \underline{107.55}
& 34.64 & \textbf{70.44} 
& 30.41 & \textbf{89.70} 
& 47.05 & \textbf{45.07} \\

\bottomrule
\end{tabular}%
}
\end{minipage}

\nextfloat

\vspace{10pt}

\begin{minipage}{\textwidth}
\centering

\begin{subfigure}[b]{0.32\linewidth}
\includegraphics[width=\linewidth]{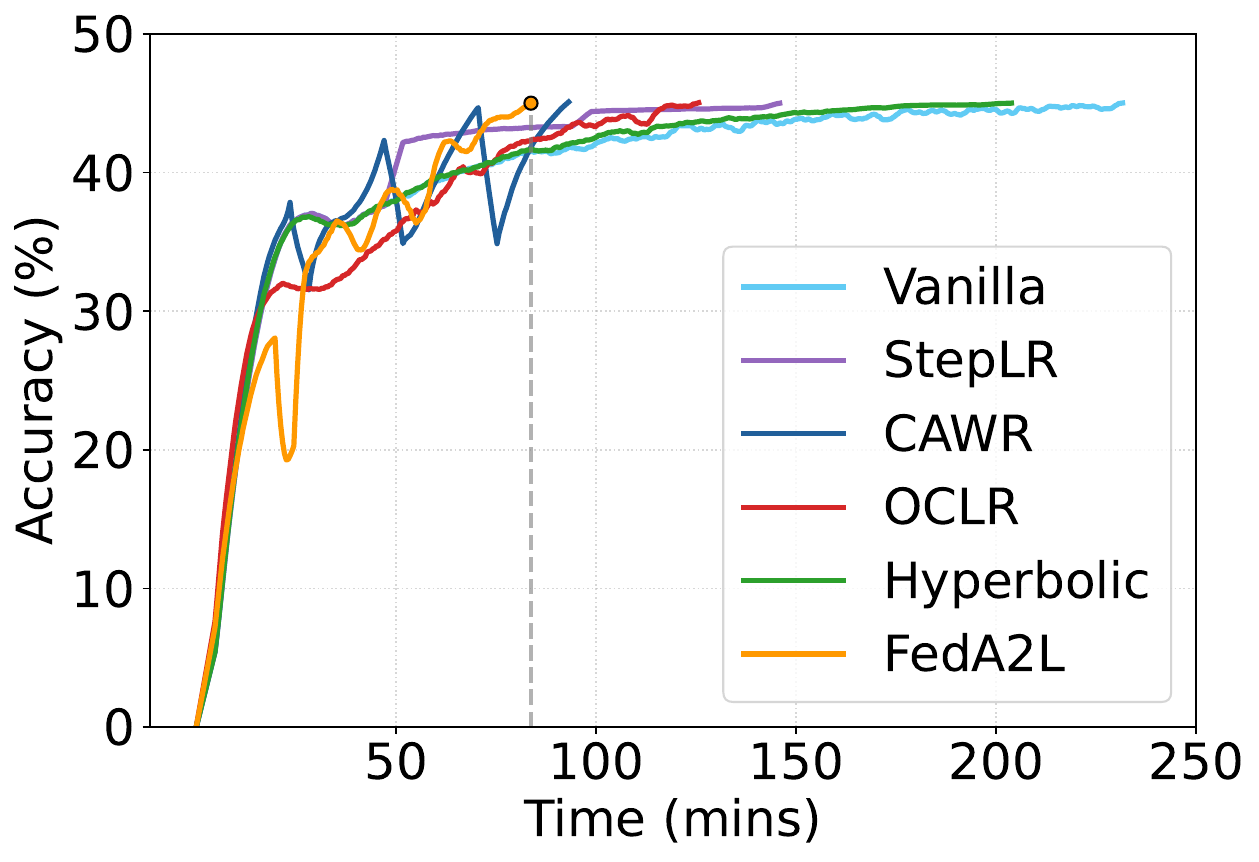}
\caption{FedAvg}
\end{subfigure}
\hfill
\begin{subfigure}[b]{0.32\linewidth}
\includegraphics[width=\linewidth]{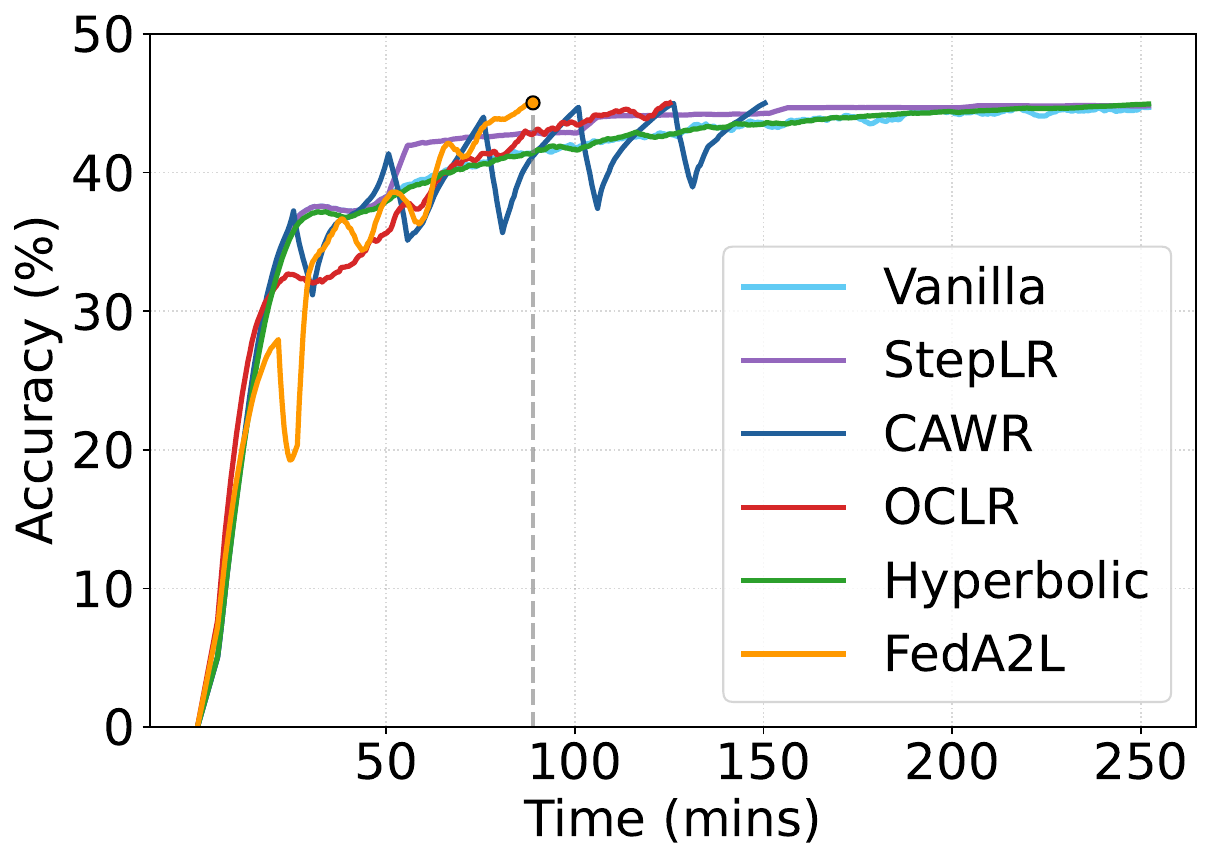}
\caption{FedProx}
\end{subfigure}
\hfill
\begin{subfigure}[b]{0.32\linewidth}
\includegraphics[width=\linewidth]{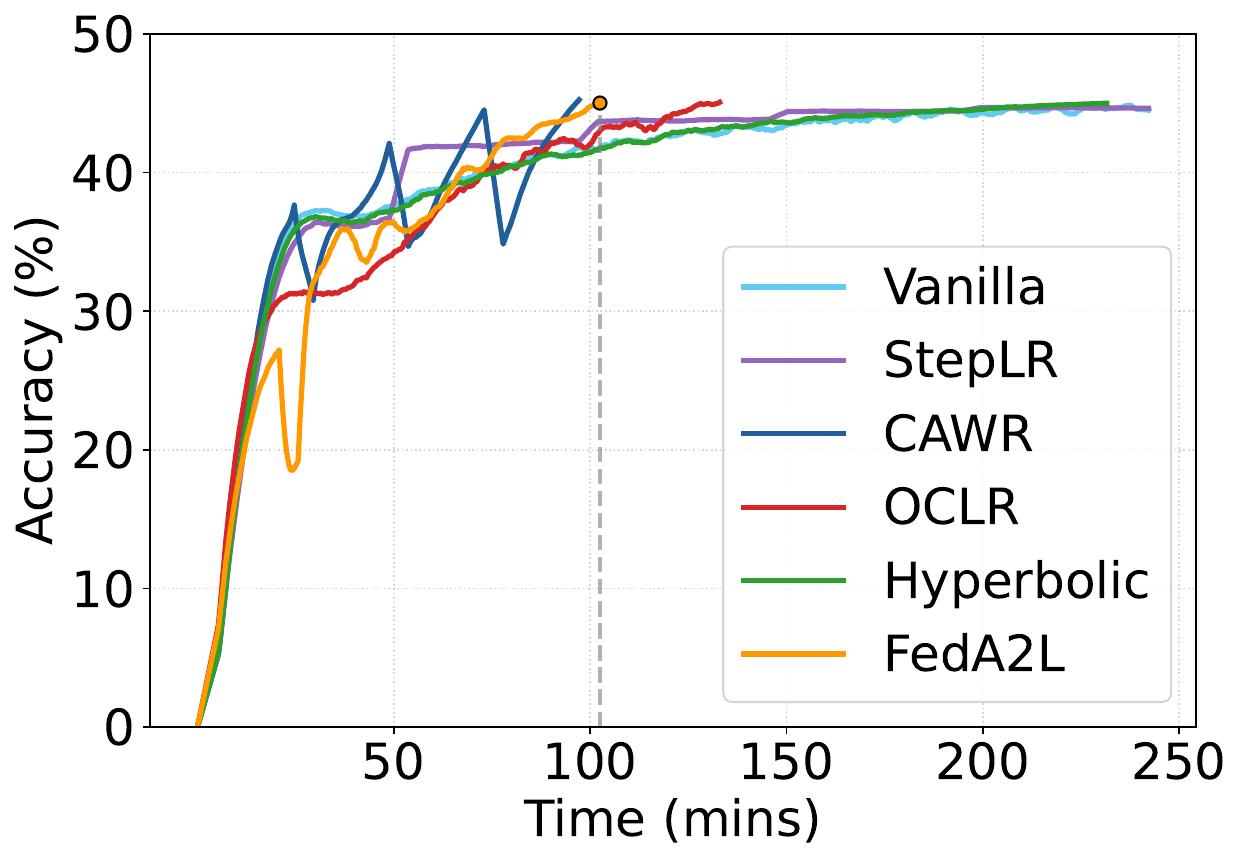}
\caption{FedYogi}
\end{subfigure}

\vspace{6pt}

\begin{subfigure}[b]{0.32\linewidth}
\includegraphics[width=\linewidth]{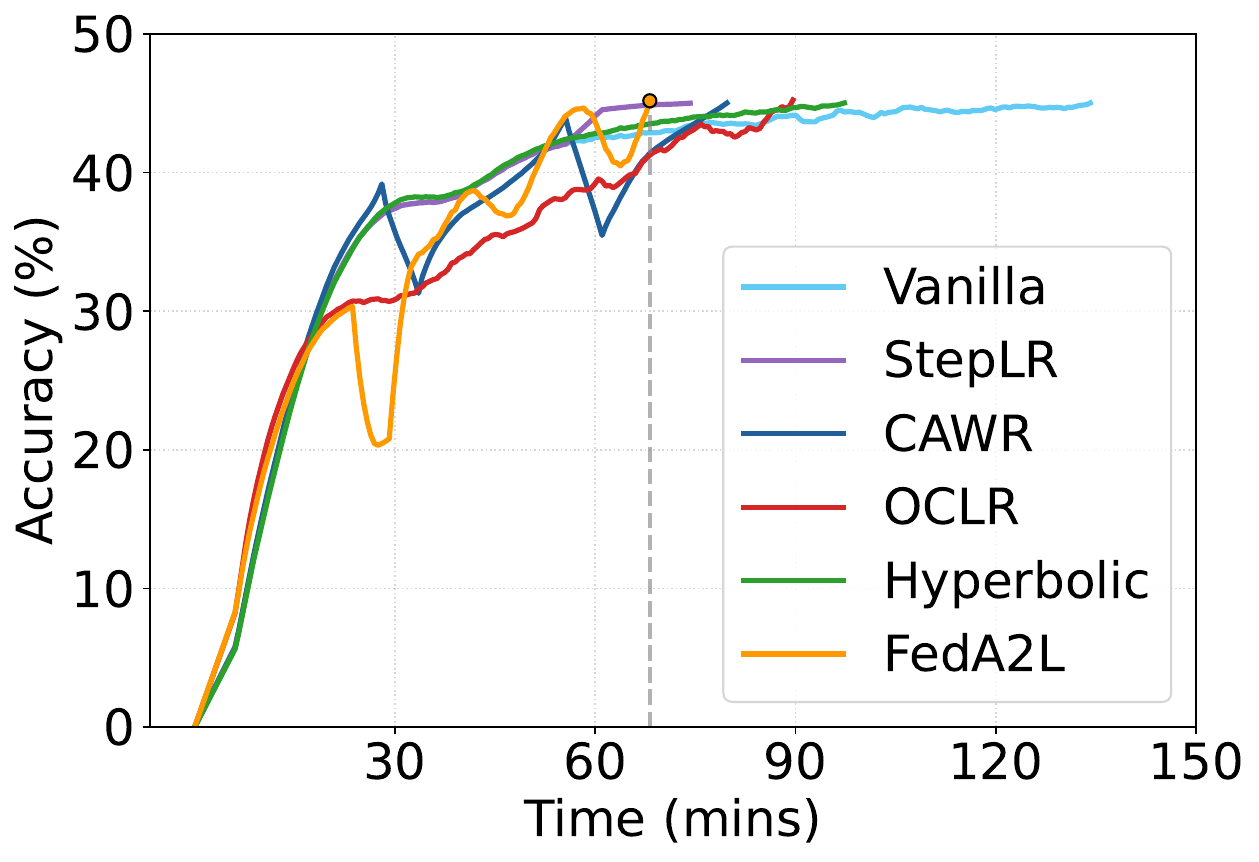}
\caption{FedNTD}
\end{subfigure}
\hfill
\begin{subfigure}[b]{0.32\linewidth}
\includegraphics[width=\linewidth]{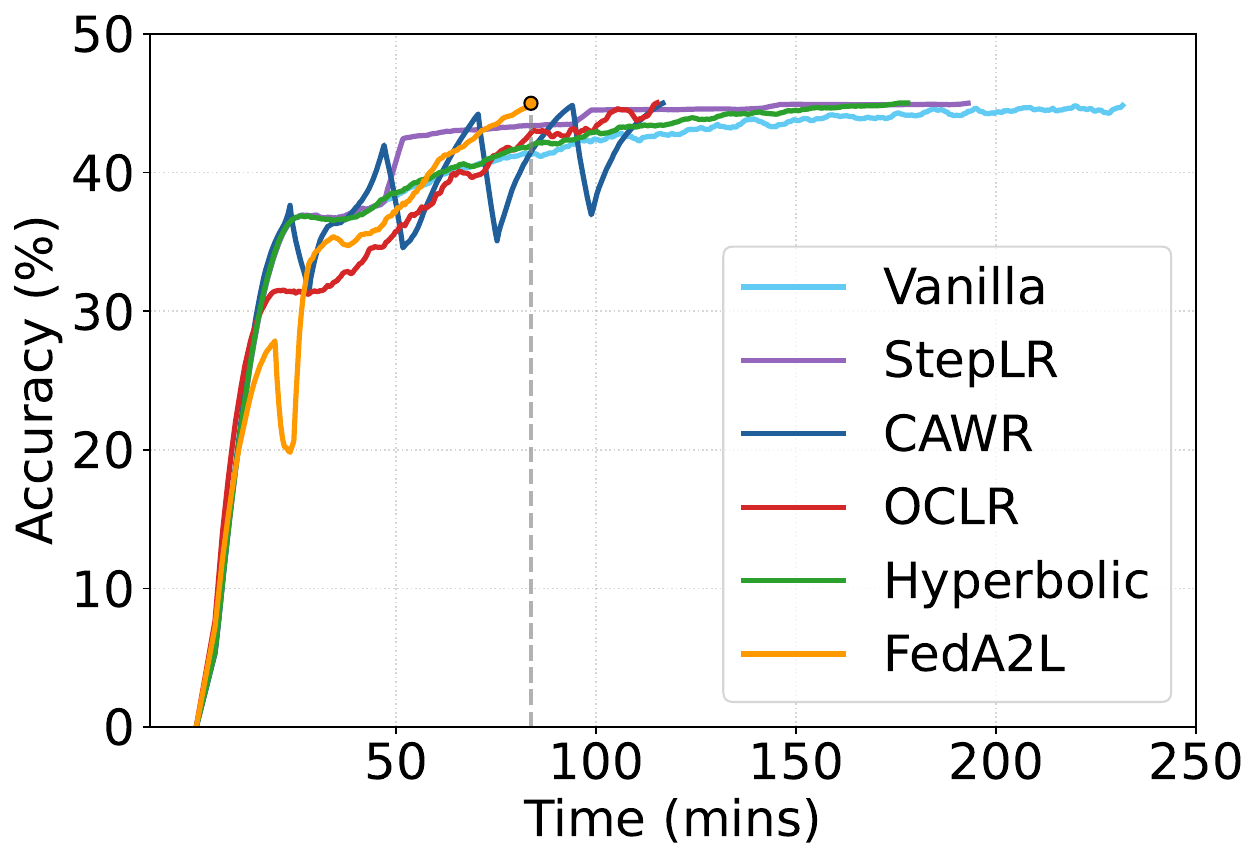}
\caption{FedAWA}
\end{subfigure}
\hfill
\begin{subfigure}[b]{0.32\linewidth}
\includegraphics[width=\linewidth]{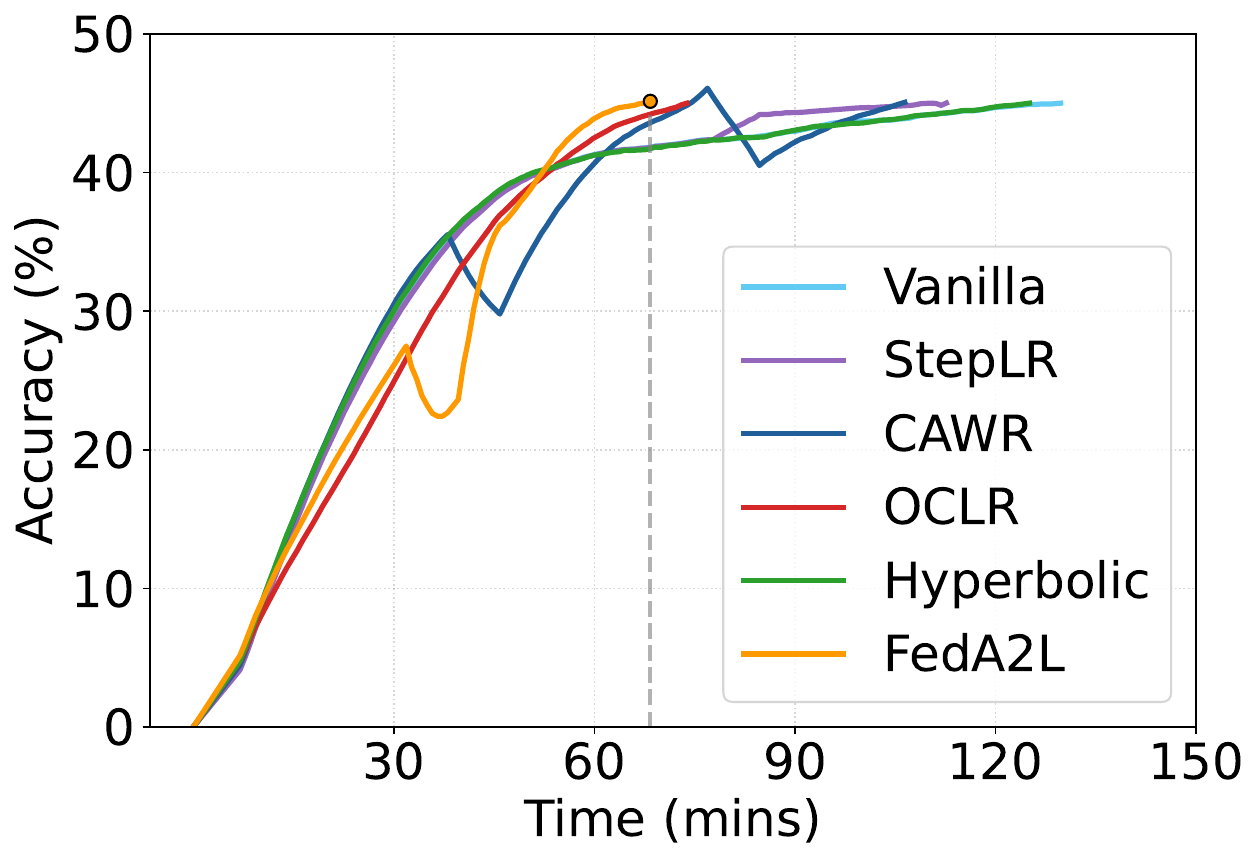}
\caption{DFedSAM}
\end{subfigure}

\end{minipage}

\vspace{6pt}

\caption{Time to reach 45\% target accuracy on CIFAR-100 using ResNet-18 (Dirichlet $\alpha=0.1$) across six DFL algorithms.}
\label{fig:convergence_time}

\end{figure*}
\subsubsection{Comparison of convergence time}

FedA2L consistently ranks within the top two in total convergence time across all six evaluated algorithms, achieving the lowest total time in five cases. As shown in Table~\ref{tab:con-time}, FedA2L incurs a modest increase in time per round relative to vanilla, for example 29.28s versus 28.43s under FedAvg, due to layer-wise metric computation. However, this overhead is consistently outweighed by the reduction in total training time, since fewer rounds are needed to reach the target accuracy. To reach the 45\% target accuracy, FedA2L converges in 177 rounds (86.38 minutes), whereas the vanilla baseline requires 491 rounds (232.63 minutes), yielding a 2.69$\times$ reduction in total time despite the higher per-round cost.

Fig.~\ref{fig:convergence_time} further illustrates this trend across all six DFL algorithms, showing that FedA2L consistently achieves the lowest total time, with the most pronounced gains under FedProx and FedAWA, where FedA2L reduces total training time by 25.1\% (91.79 vs.\ 122.60 minutes) and 21.65\% (89.70 vs.\ 114.48 minutes) relative to the best competing baseline, respectively (Table~\ref{tab:con-time}). These results demonstrate that FedA2L’s layer-wise adaptation provides measurable wall-clock efficiency across diverse DFL protocols, supporting its practicality for time-sensitive industrial IoT, cyber-physical systems, and decentralized intelligence deployments where both communication and computational costs are critical.

\begin{figure*}[!th]
    \centering
    \begin{subfigure}[b]{0.32\linewidth}
        \centering
        \includegraphics[width=\linewidth]{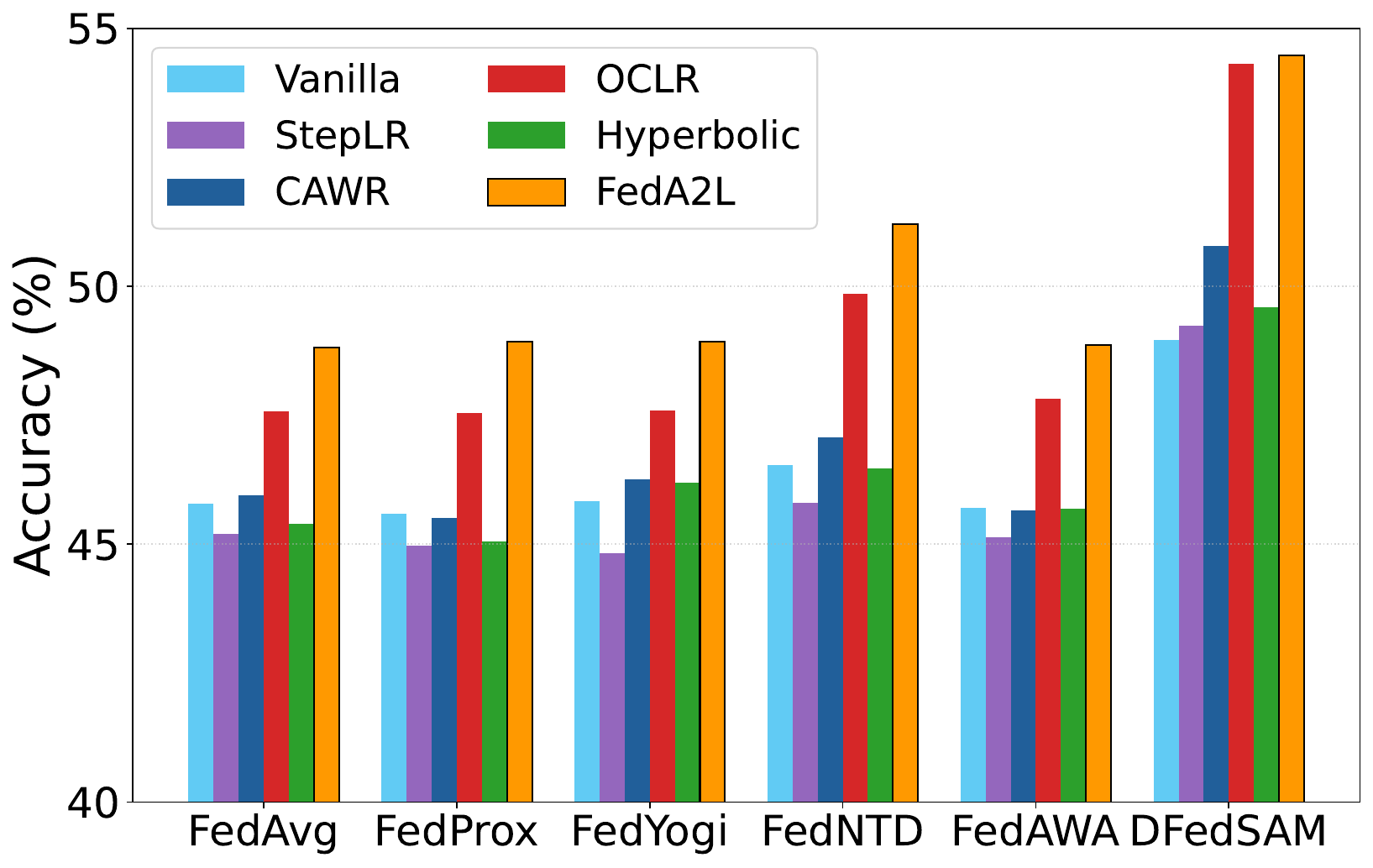}
        \caption{ResNet-18 on CIFAR-100}
        \label{fig:convergence_fedavg}
    \end{subfigure}
    \hfill
    \begin{subfigure}[b]{0.32\linewidth}
        \centering
        \includegraphics[width=\linewidth]{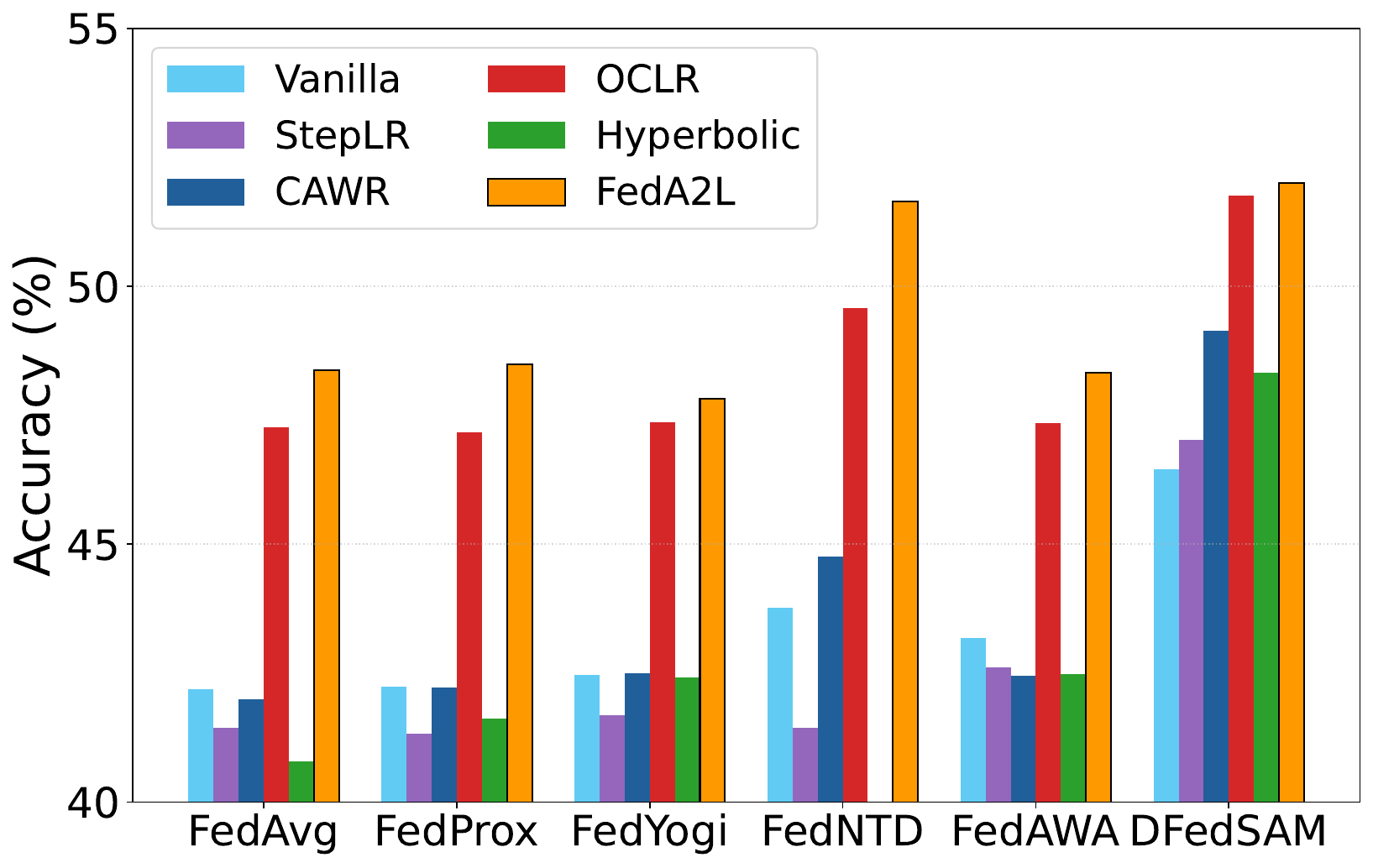}
        \caption{ResNet-34 on CIFAR-100}
        \label{fig:convergence_fedprox}
    \end{subfigure}
    \hfill
    \begin{subfigure}[b]{0.32\linewidth}
        \centering
        \includegraphics[width=\linewidth]{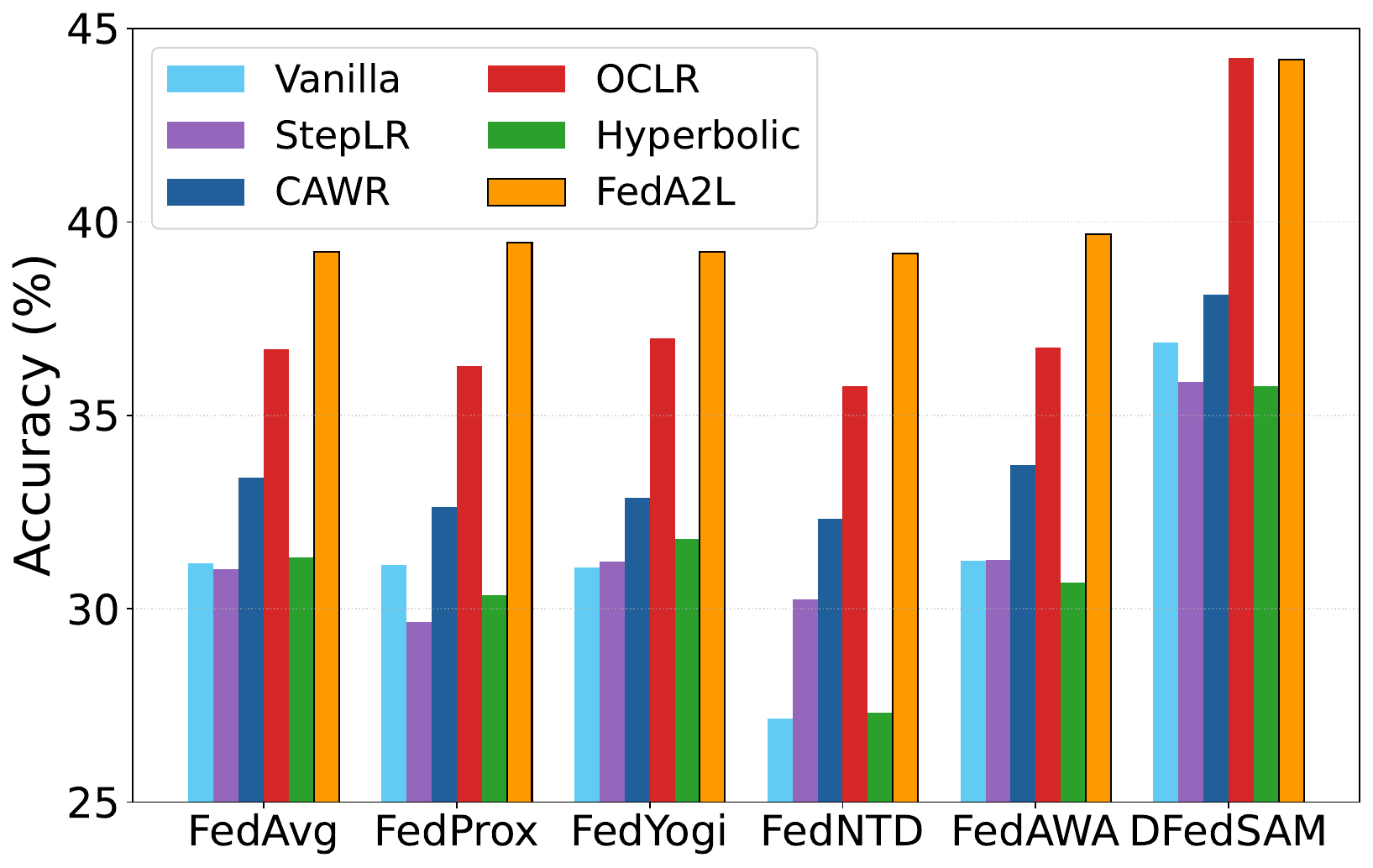}
        \caption{ResNet-34 on TinyImageNet}
        \label{fig:convergence_fedyogi}
    \end{subfigure}
    \caption{Highest test accuracy achieved across algorithms, models, and datasets under non-IID partitioning (Dirichlet $\alpha=0.1$).}
    \label{fig:robustacc}
\end{figure*}
\subsection{How well does FedA2L perform in accuracy?}

Beyond accelerating convergence, FedA2L achieves the highest test accuracy across diverse settings, maintaining both model quality and training stability under non-IID conditions. Fig.~\ref{fig:robustacc} compares the highest achieved accuracy as model depth and dataset complexity increase. FedA2L reaches the highest accuracy in the majority of configurations across all six algorithms. The accuracy improvement stems from the same underlying mechanism as convergence improvement, by dynamically balancing layer-specific roles based on local divergence metrics. FedA2L avoids the trade-off inherent in uniform LRs where high rates destabilize and low rates under-adapt, maintaining both leading accuracy and stable convergence across model architectures and datasets. Overall, fine-grained per-layer control contributes to consistently higher accuracy without compromising training stability in heterogeneous decentralized environments.

\subsection{How robust is FedA2L in practical DFL settings?}
\begin{table*}
\captionsetup{
  width=0.8\textwidth,
  justification=raggedright,
  singlelinecheck=false
}
\centering
\caption{Convergence performance of FedA2L vs. baselines in DFL frameworks. Rounds to target accuracies varying data heterogeneity, scalability, and network topologies. Results are smoothed using a sliding average (window size $=15$). Formatting: Bold = fastest; Underline = second-fastest; Dash = failure to converge.}
\resizebox{0.8\linewidth}{!}{
\begin{tabular}{cc|cc|cc|cc}
\toprule
 \multicolumn{2}{c|}{Setting} & \multicolumn{2}{c|}{Heterogeneity} & \multicolumn{2}{c|}{Scalability} & \multicolumn{2}{c}{Topology} \\
 \cmidrule(lr){1-2} 
 \cmidrule(lr){3-4} 
 \cmidrule(lr){5-6} 
 \cmidrule(lr){7-8} 
 \multicolumn{2}{c|}{Target Accuracy} & 40\% & 45\% & 42\% & 40\% & 33\% & 45\% \\
 \cmidrule(lr){1-2} 
 \cmidrule(lr){3-4} 
 \cmidrule(lr){5-6} 
 \cmidrule(lr){7-8} 
Algorithm & Method & \textit{Dir}(0.01) &\textit{Dir}(0.5) & 20 nodes & 30 nodes & Ring & K-Connected\\
\midrule
\multirow{6}{*}{FedAvg}
    & Vanilla    &300  &354  &\_   &\_  &\_  &\_  \\
    & StepLR     &206  &\underline{109}  &\_   &\_  &\_  &\_  \\
    & CAWR       &\underline{194}  &136  &398  &\underline{348}  &\_  &\_  \\
    & OCLR       &298  &132  &\underline{341}  &366  &\underline{493}  &\underline{296}  \\
    & Hyperbolic &315  &248  &\_   &\_  &\_  &\_  \\
    & FedA2L     &\textbf{181}     &\textbf{104}  &\textbf{278}  &\textbf{339}  &\textbf{318}  &\textbf{216}  \\
\midrule
\multirow{5}{*}{FedAWA}
    & Vanilla    &380   &406  &\_  &\_  &\_  &\_  \\
    & StepLR     &209  &\underline{125}  &\_  &\_  &\_  &\_  \\
    & CAWR       &\underline{193}  &140  &\_  &451  &\_  &299  \\
    & OCLR       &316  &135  &\underline{352}  &\underline{379}  &\underline{497}  &\underline{296}  \\
    & Hyperbolic &314  &226  &\_  &\_  &\_  &\_  \\
    & FedA2L     &\textbf{187}  &\textbf{106}  &\textbf{273}  &\textbf{334}  &\textbf{319}  &\textbf{211}  \\
\bottomrule
\end{tabular}}
\label{tab:robustness}
\end{table*}

To validate practical applicability beyond baseline conditions, this section evaluates FedA2L's robustness under severe non-IID data distributions, increased node scalability, and challenging network topologies (Table~\ref{tab:robustness}). Across these demanding conditions, FedA2L consistently reduces communication rounds required to reach target accuracy, demonstrating that adaptive layer-wise LRs remain effective and stable as DFL training conditions become more challenging.

\subsubsection{Comparison under heterogeneous data distributions} 

The convergence results under heterogeneous data distributions (Table~\ref{tab:robustness}, heterogeneity columns) underscore the robust capability of FedA2L to mitigate the severe impact of non-IID data. Under the highly skewed $\alpha=0.01$ conditions that create extreme optimization challenges, baseline performance degrades sharply due to client drift where inconsistent local gradients pull the model away from the optimum. In contrast, FedA2L achieves the fastest convergence (181 rounds for FedAvg, 187 for FedAWA) and maintains a substantial $1.66\times$ speedup over the vanilla baseline. This result validates that the layer-wise adaptive mechanism remains effective, enabling a stable and accelerated convergence that is crucial in practical non-IID settings.

\subsubsection{Comparison across node scalability}

The evaluation of node scalability (Table~\ref{tab:robustness}, scalability columns), from 20 to 30 clients, confirms that the performance benefits of FedA2L are consistent in increasingly complex DFL environments, with more distributed interactions, even as the consensus challenge grows. Larger networks introduce more gradients to aggregate, amplifying noise and slowing consensus propagation. FedA2L sustains its performance advantage over all baselines, requiring 278 and 273 rounds under FedAvg and FedAWA, respectively, for 20 nodes, and 339 and 334 rounds for 30 nodes. These results demonstrate that FedA2L effectively mitigates this increased noise and the ensuing consensus delays inherent in larger peer-to-peer networks, allowing layer-wise adaptive LRs to remain effective without proportional degradation as network size increases. 

\subsubsection{Comparison across network topologies}

The evaluation across network topologies (Table~\ref{tab:robustness}, topology columns), particularly the challenging sparse Ring topology, provides compelling evidence of FedA2L's resilience to communication bottlenecks. The Ring topology is structurally demanding due to its high diameter and severely restricted neighbor interactions, causing most baseline methods to fail to converge entirely. Despite these severe structural constraints, FedA2L successfully converges, achieving a remarkable $1.55\times$ speedup (318 rounds for FedAvg, 319 for FedAWA) compared to the best-performing converging baseline (OCLR). This demonstrates that layer-wise adaptive LRs effectively compensate for structural communication limitations, enabling faster convergence even under severely constrained neighbor interactions.

\subsection{Results on TSF}
Table~\ref{tab:timeseries} reports communication rounds and total time to reach target MSE on both datasets under FedAvg and FedAWA. FedA2L achieves the fastest convergence across all settings. On ExchangeRate, FedA2L reaches target MSE in 11 rounds versus 16 rounds for the best scheduler (OCLR) under FedAvg, a $1.45\times$ speedup that translates to 31.25\% reduction in communication rounds and 26.37\% reduction in total time. Under FedAWA, the gap widens: 11 versus 17 rounds ($1.55\times$ speedup, 35.29\% fewer rounds, 31.67\% less time). On BeijingAirQuality, the improvement is more modest: 18 versus 22 rounds under FedAvg (13.64\% fewer rounds, 9.26\% less time) and 20 versus 22 rounds under FedAWA (9.1\% fewer rounds, 5.91\% less time). The smaller gains indicate that highly multivariate, high-frequency data (11 variates, hourly) in BeijingAirQuality exhibits less layer-wise optimization conflict than univariate, low-frequency data (1 variate, daily) in ExchangeRate, where per-layer LR tuning has more room to improve. In all cases, FedA2L's modest per-round overhead (28-31 s versus 28-29 s for the baselines) is outweighed by the reduction in total rounds required, confirming the practical benefit of adaptive layer-wise LRs across both temporal forecasting settings.

\begin{table}[t]
\centering
\captionsetup{
  width=0.9\columnwidth,
  justification=raggedright,
  singlelinecheck=false
}
\caption{Communication rounds and total time to reach target MSE (0.0575 for ExchangeRate, 4.3582 for BeijingAirQuality) for FedA2L and baseline learning rate strategies.}
\label{tab:timeseries}
\resizebox{\columnwidth}{!}{
\begin{tabular}{cc|cc|cc}
\toprule
\multicolumn{2}{c|}{Setting} 
& \multicolumn{2}{c|}{ExchangeRate} 
& \multicolumn{2}{c}{BeijingAirQuality} \\
\cmidrule(lr){1-2}
\cmidrule(lr){3-4}
\cmidrule(lr){5-6}

Algorithm & Method
& Round & \begin{tabular}[c]{@{}c@{}}Total time \\ (s)\end{tabular}
& Round & \begin{tabular}[c]{@{}c@{}}Total time \\ (s)\end{tabular} \\
\midrule

\multirow{6}{*}{FedAvg}
& Vanilla    &47  &1170.46  &48  &1195.36  \\
& StepLR     &55  &1365.63  &\_  &\_  \\
& CAWR       &89  &2180.64  &38  &1376.40  \\
& OCLR       &\underline{16}  &\underline{393.86}   &\underline{22}  &\underline{806.05}  \\
& Hyperbolic &88  &2158.42  &42  &1519.98  \\
& FedA2L     &\textbf{11}  &\textbf{289.98 }  &\textbf{18}  &\textbf{731.38}  \\
& \textit{Improv. }   &\textit{31.25\%}  &\textit{26.37\%}  &\textit{13.64\%}    &\textit{9.26\%} \\
\midrule

\multirow{6}{*}{FedAWA}     
& Vanilla    &58  &1391.18  &40  &1547.33  \\
& StepLR     &\_  &2121.43  &\_  &\_ \\
& CAWR       &98  &2360.4   &34  &1307.71  \\
& OCLR       &\underline{17}  &\underline{423.39}   &\underline{22}  &\underline{852.10}  \\
& Hyperbolic &\_  &2379.8   &38  &1470.10  \\
& FedA2L     &\textbf{11 } &\textbf{307.31}   &\textbf{20}  &\textbf{801.69 } \\
& \textit{Improv.  }  &\textit{35.29\%} &\textit{31.67\%} &\textit{9.1\%} &\textit{5.91\%} \\
\bottomrule
\end{tabular}}
\end{table}

\subsection{Hyperparameter sensitivity analysis and ablation study} \label{sec:ablation}
We conduct a two-part analysis to examine the design choices and configuration parameters of FedA2L. First, we analyze the dynamics and stability of the layer-wise LR feedback mechanism to verify that the adaptive behavior is driven by meaningful optimization signals rather than noise. Second, we conduct a hyperparameter sensitivity and component ablation analysis to evaluate how the design parameters and the dual-metric structure affect both convergence speed and model performance.

\subsubsection{LR feedback dynamics and stability}

\begin{figure*}
    \centering
    \begin{subfigure}[b]{0.32\linewidth}
        \centering
        \includegraphics[width=\linewidth]{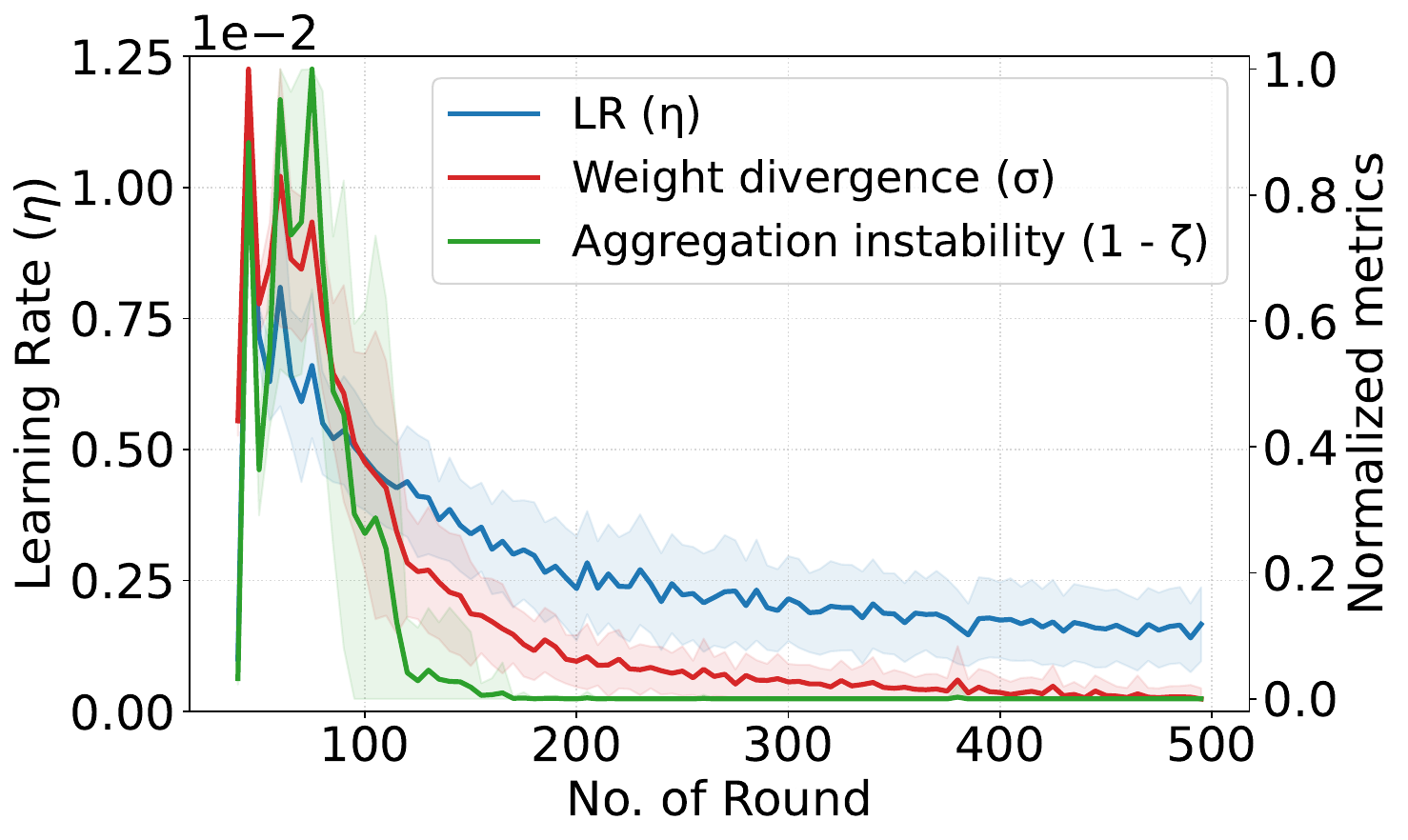}
        \caption{Highest divergence layer}
    \end{subfigure}
    \hfill
    \begin{subfigure}[b]{0.32\linewidth}
        \centering
        \includegraphics[width=\linewidth]{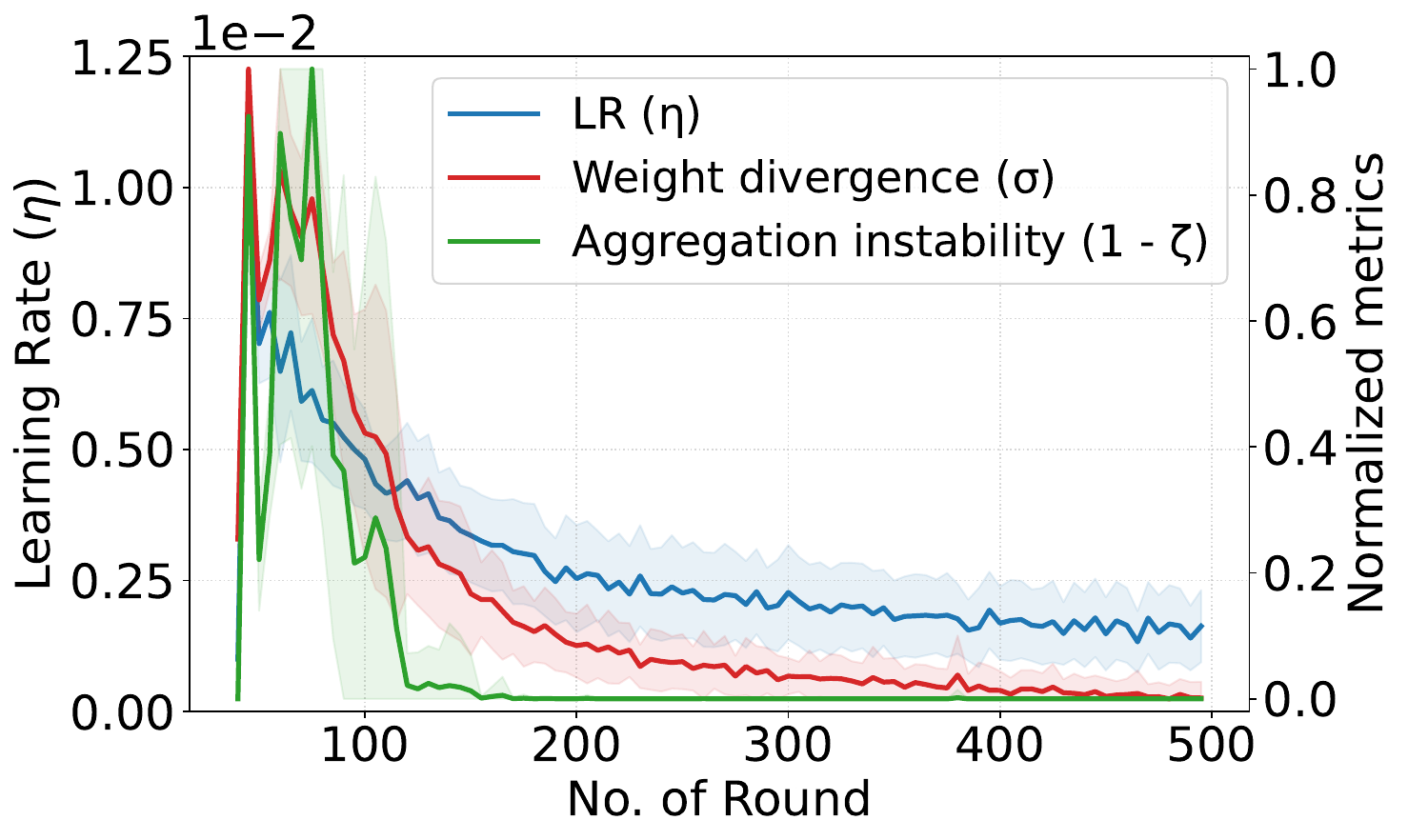}
        \caption{Second highest divergence layer}
    \end{subfigure}
    \hfill
    \begin{subfigure}[b]{0.32\linewidth}
        \centering
        \includegraphics[width=\linewidth]{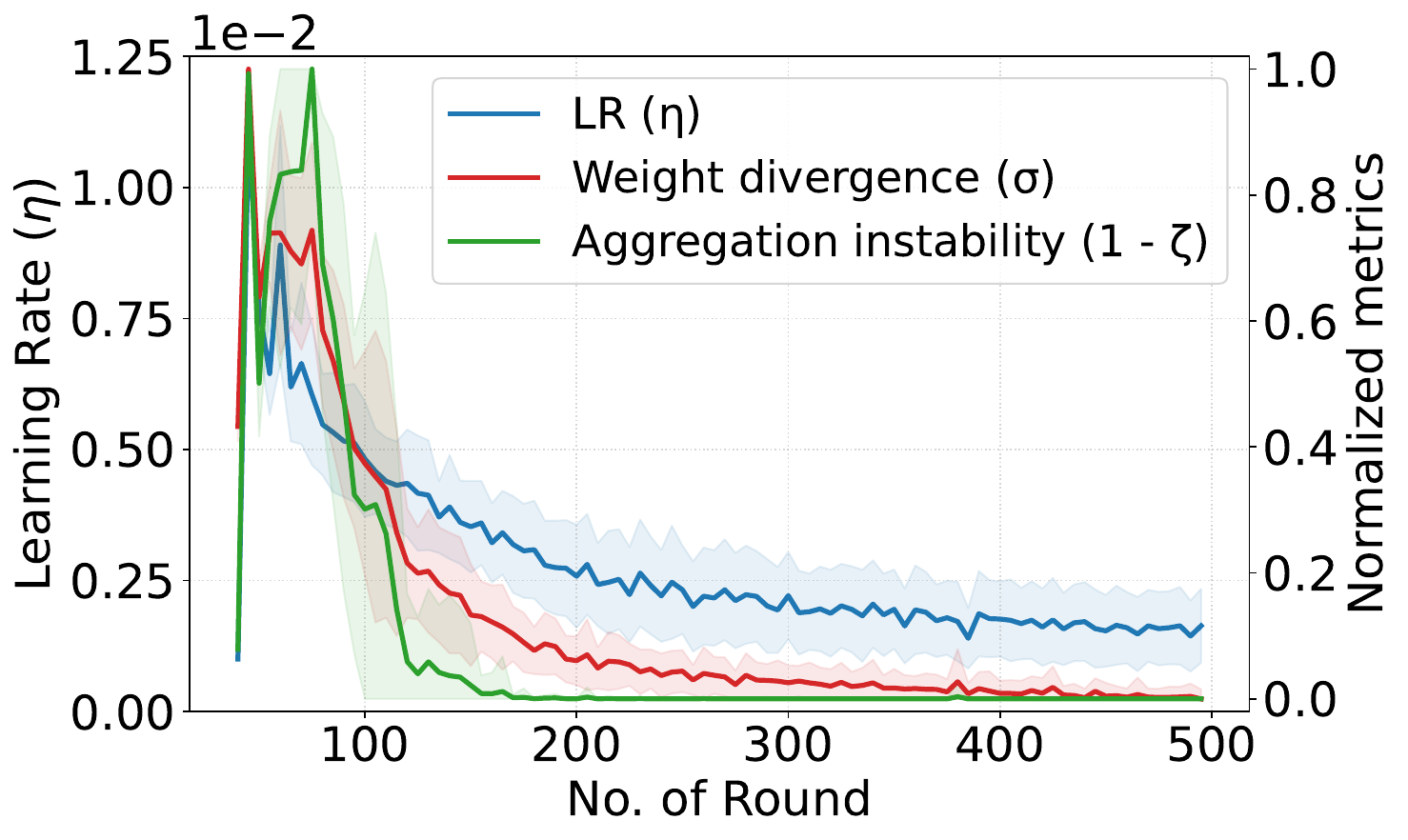}
        \caption{Third highest divergence layer}
    \end{subfigure}
    \caption{LR feedback dynamics for the three layers with the highest local divergence (ResNet-18, CIFAR-100, FedAvg, Dirichlet $\alpha=0.1$). The plots show the layer-wise learning rate $\eta_{i,l}^r$ (left axis), together with the normalized weight divergence $\sigma_{i,l}^r$ and aggregation instability $1-\zeta_{i,l}^r$ (right axis), over communication rounds.}
    \label{fig:lr_visualization}

\end{figure*}
To inspect the behavior of FedA2L, we analyze the joint evolution of the layer-wise LR and the local divergence metrics. Fig.~\ref{fig:lr_visualization} shows the top three layers ranked by their highest local divergence metrics in a representative configuration (ResNet-18, CIFAR-100, FedAvg at $\alpha=0.1$). For each selected layer, we plot across rounds the effective LR $\eta_{i,l}^r$ together with the normalized weight divergence $\sigma_{i,l}^r$ and aggregation instability $1-\zeta_{i,l}^r$\footnote{We use $1-\zeta_{i,l}^r$ instead of aggregation stability ($\zeta_{i,l}^r$) so that higher values consistently indicate stronger neighbor corrections, aligning the direction of both instability signals. All metric curves are linearly rescaled to $[0,1]$ for visualization.}.

Across all three layers, we observe large spikes in both $\sigma_{i,l}^r$ and $1-\zeta_{i,l}^r$ during early rounds, indicating strong local updates and non-negligible corrections by neighbors. In response, the corresponding LRs temporarily increase but remain within a narrow range and then decay smoothly over time. As training progresses, the divergence and aggregation-instability signals rapidly diminish toward zero, and $\eta_{i,l}^r$ converges toward a stable plateau determined by the global decay factor $\gamma^r$. Importantly, even when a layer exhibits large divergence or strong correction, the LR does not grow without bound; transient boosts are followed by reductions, consistent with the bounded $1+\tanh(\log(\lambda_{i,l}^r))$ modulation and the self-correcting Z-score normalization described in Section~\ref{sec:fedA2L_lr}.

These trajectories show that FedA2L briefly increases LRs when local signals are strong, enabling short-term beneficial divergence in specialized layers under non-IID data, while the normalization and decay mechanisms gradually return the rates toward the global schedule and prevent runaway behavior.

\begin{table*}[t]
\centering
\caption{Hyperparameter sensitivity and ablation analysis of FedA2L on CIFAR-100 ($\alpha=0.1$). Results are reported in terms of communication rounds to reach the target accuracy and best test accuracy (\%).}

\label{tab:fedA2L_sensitivity}

\resizebox{\linewidth}{!}{%
\begin{tabular}{cc|ccc|ccc|ccc|ccc|ccc}
\toprule

\multirow{2}{*}{Model}
& \multirow{2}{*}{Metric}
& \multicolumn{3}{c|}{$\tau$}
& \multicolumn{3}{c|}{$R_{\text{warm}}$}
& \multicolumn{3}{c|}{$\rho$}
& \multicolumn{3}{c|}{$\xi$}
& \multicolumn{3}{c}{$\beta$}
\\

\cmidrule(lr){3-5}
\cmidrule(lr){6-8}
\cmidrule(lr){9-11}
\cmidrule(lr){12-14}
\cmidrule(lr){15-17}

&
& $10^{-2}$ & $10^{-3}$ & $10^{-4}$
& 10 & 20 & 40
& 5 & 10 & 20
& 0.05 & 0.1 & 0.3
& 0.0 & 0.6 & 1.0
\\

\midrule

\multirow{2}{*}{CNN}
& Rounds@29\%
& 22 & \textbf{21} & 23
& \textbf{24} & 38 & 29
& \textbf{23} & 28 & 25
& 30 & 29 & \textbf{24}
& 23 & \textbf{21} & 23
\\

& Best accuracy
& \textbf{33.39} & \textbf{33.39} & 33.31
& \textbf{33.11} & 32.59 & 31.14
& \textbf{33.29} & 32.97 & 32.88
& 32.33 & 32.87 & \textbf{33.33}
& 32.58 & \textbf{33.63} & 32.46
\\

\midrule

\multirow{2}{*}{ResNet-18}
& Rounds@45\%
& 173 & \textbf{171} & 192
& 462 & 207 & \textbf{175}
& 197 & \textbf{176} & 179
& 179 & \textbf{177} & \_
& 176 & \textbf{175} & 180
\\

& Best accuracy
& 48.46 & \textbf{48.81} & 48.53
& 45.46 & 47.78 & \textbf{48.63}
& 46.94 & \textbf{48.78} & 48.69
& 48.73 & \textbf{48.86} & 44.90
& 49.23 & \textbf{49.50} & 49.10
\\

\midrule

\multirow{2}{*}{ResNet-34}
& Rounds@45\%
& 148 & \textbf{146} & 153
& 249 & 167 & \textbf{156}
& 160 & \textbf{156} & \textbf{156}
& 168 & \textbf{157} & \_
& 158 & \textbf{150} & 159
\\

& Best accuracy
& \textbf{48.67} & \textbf{48.67} & 48.61
& 46.52 & 48.63 & \textbf{48.87}
& 48.27 & 48.78 & \textbf{48.86}
& 48.73 & \textbf{48.86} & 44.29
& 47.22 & \textbf{48.63} & 47.22
\\

\bottomrule
\end{tabular}%
}
\end{table*}

\subsubsection{Hyperparameter sensitivity and configuration} \label{sec:hyperparam}

\begin{figure*}
    \centering
    \begin{subfigure}[b]{0.32\linewidth}
        \centering
        \includegraphics[width=\linewidth]{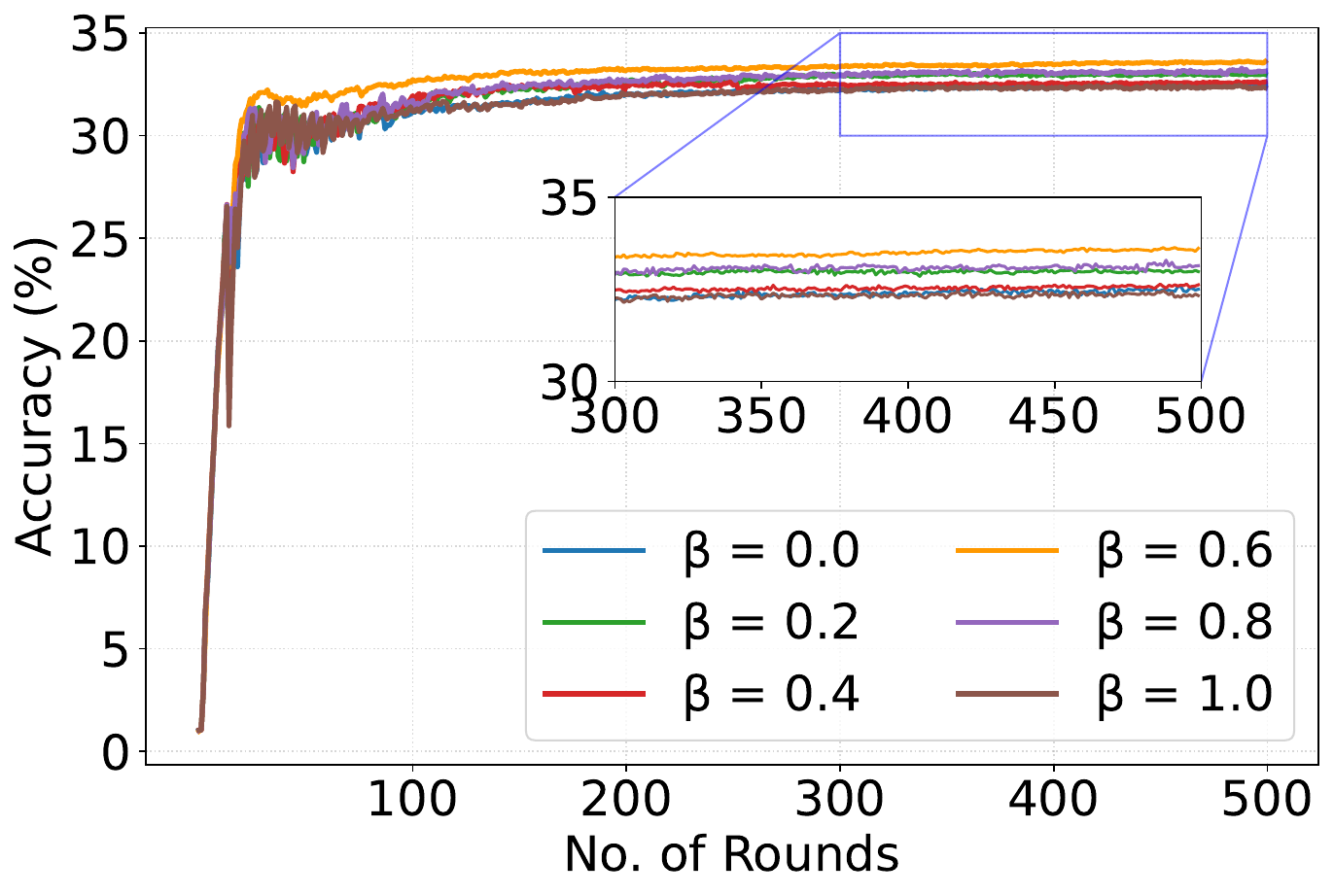}
        \caption{CNN}
        \label{fig:beta_cnn}
    \end{subfigure}
    \hfill
    \begin{subfigure}[b]{0.32\linewidth}
        \centering
        \includegraphics[width=\linewidth]{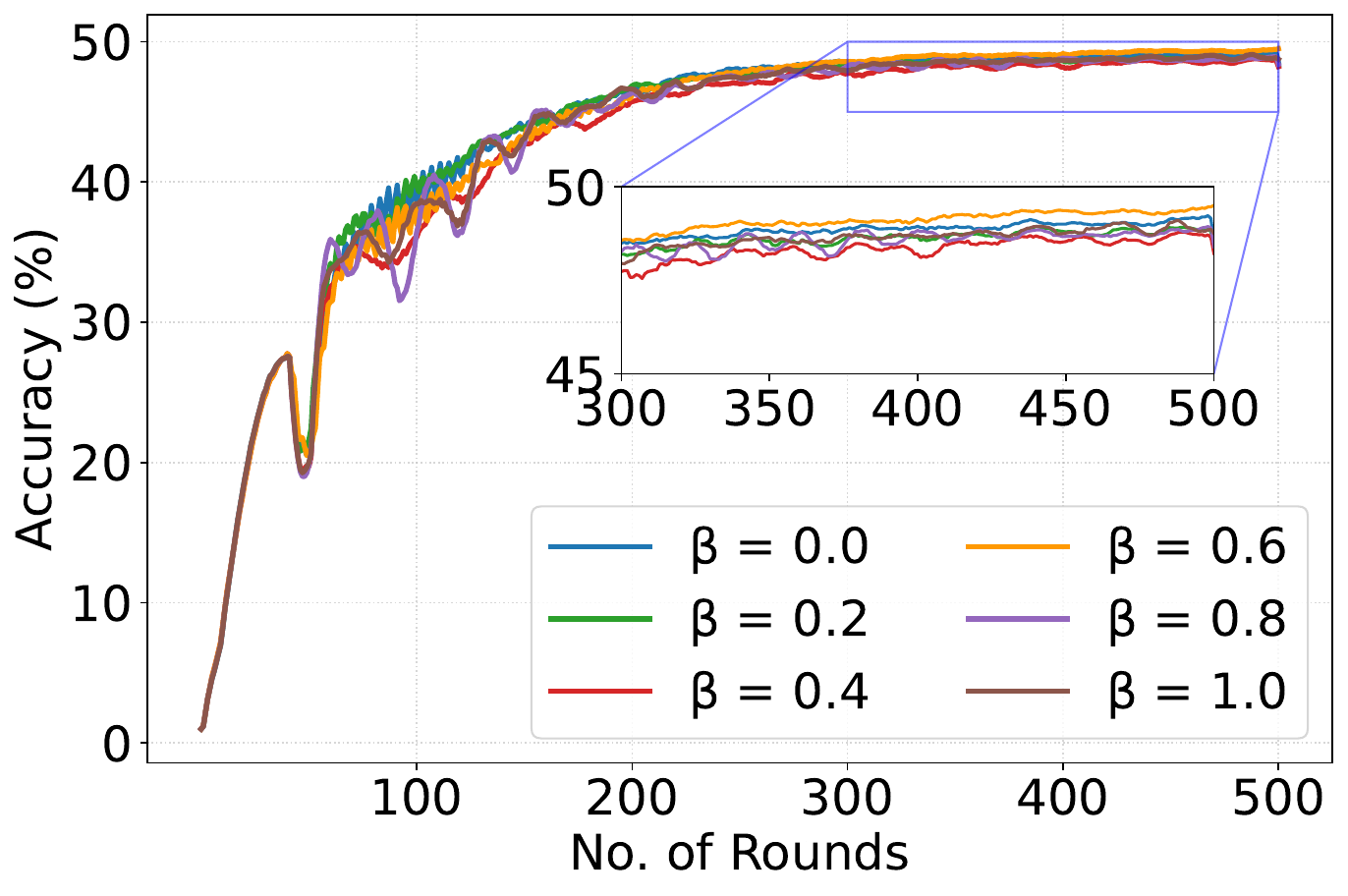}
        \caption{ResNet-18}
        \label{fig:beta_resnet18}
    \end{subfigure}
    \hfill
    \begin{subfigure}[b]{0.32\linewidth}
        \centering
        \includegraphics[width=\linewidth]{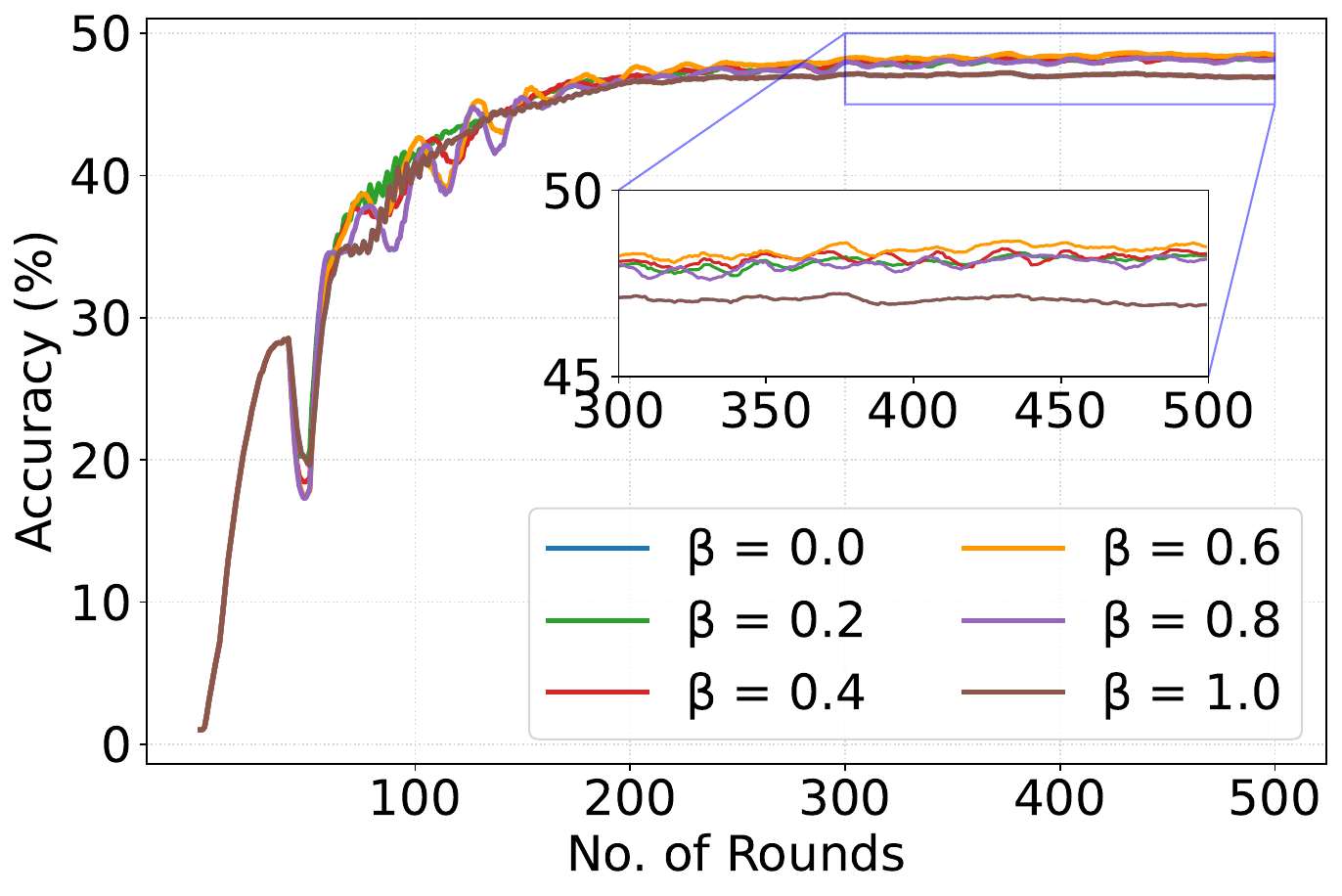}
        \caption{ResNet-34}
        \label{fig:beta_resnet34}
    \end{subfigure}
    \caption{Effect of fusion balance $\beta$ on convergence speed and model performance under FedAvg on CIFAR-100. The extreme values $\beta = 0$ ($\zeta$-only) and $\beta = 1$ ($\sigma$-only) constitute direct component ablations of $\zeta$ and $\sigma$, respectively, in Eq.~\ref{eq:fusion}, while intermediate values reflect the full dual-metric fusion.}
    \label{fig:beta}

\end{figure*}

To assess robustness and reproducibility, we conduct a sensitivity study on CIFAR-100 across CNN (29\% target), ResNet-18 and ResNet-34 (45\% target), varying each of the five FedA2L design parameters ($R_{\text{warm}}$, $\rho$, $\xi$, $\tau$, and $\beta$) independently while holding the others at their defaults. Table~\ref{tab:fedA2L_sensitivity} reports both convergence efficiency and best test accuracy for each parameter, providing a joint view of how each parameter affects both dimensions of performance. Fig.~\ref{fig:beta} presents the full accuracy trajectories under varying
$\beta$, where the extreme settings $\beta=0$ and $\beta=1$ serve as direct ablations of the individual $\zeta$ and $\sigma$ components, respectively, enabling assessment of their contribution to both convergence behavior and final model accuracy.

\textbf{Influence of stability threshold ($\tau$)}. The stability threshold $\tau$ in Eq.~\ref{eq:zeta_stability} sets the tolerance boundary for counting a parameter as stable during aggregation, directly controlling the resolution of the consensus signal $\zeta$. Too small a value makes $\zeta$ uninformative by classifying nearly all parameter changes as unstable; too large a value loses discrimination between genuine consensus and client drift. As shown in Table~\ref{tab:fedA2L_sensitivity}, CNN is negligibly affected across all tested values, while $\tau = 10^{-4}$ degrades convergence on ResNet-18 and ResNet-34 from 171 to 192 rounds and from 146 to 153 rounds respectively, confirming that $\zeta$ loses its discriminative power at excessively fine thresholds. A stable operating region exists between $10^{-2}$ and $10^{-3}$, and $\tau = 10^{-3}$ achieves competitive accuracy across all architectures, confirming that convergence efficiency and model quality are jointly optimized. We adopt $\tau = 10^{-3}$ across all model architectures.

\textbf{Influence of warm-up rounds ($R_{\text{warm}}$)}. The warm-up period $R_{\text{warm}}$ defines the number of initial rounds during which FedA2L applies the base LR uniformly across all layers before adaptive modulation begins. Its sensitivity scales with model depth, where shallow architectures accumulate reliable divergence statistics within a few rounds, whereas deeper models require a longer history for per-layer signals to stabilize. As shown in Table~\ref{tab:fedA2L_sensitivity}, reducing $R_{\text{warm}}$ to 10 rounds degrades convergence from 175 to 462 rounds on ResNet-18 and from 156 to 249 rounds on ResNet-34, while all CNN settings yield comparable results. Critically, very short warm-up periods also reduce best accuracy, as insufficient metric history produces unreliable early LR modulation that degrades the full training trajectory. We adopt $R_{\text{warm}} = 10$ for CNN and $R_{\text{warm}} = 40$ for ResNet architectures.

\textbf{Influence of temporal window size ($\rho$)}. The window size $\rho$ controls the number of recent rounds used in Z-score normalization, directly governing the temporal horizon of the adaptive signal. FedA2L is largely insensitive to this parameter because Z-score normalization inherently reduces short-term fluctuations regardless of window length, making the adaptive signal robust to the exact history size. Table~\ref{tab:fedA2L_sensitivity} confirms this robustness: convergence varies within a narrow range across $\rho \in \{5, 10, 20\}$ on all architectures, ranging from 23 to 25 rounds on CNN and from 176 to 197 rounds on ResNet-18, without a consistent directional trend, and best accuracy follows the same pattern across all architectures. ResNet architectures show marginally better convergence with $\rho = 10$, consistent with their greater need for stable statistical history. We adopt $\rho = 5$ for CNN and $\rho = 10$ for ResNet architectures as minimal window sizes that ensure sufficient statistical stability.

\textbf{Influence of global decay constant ($\xi$)}. The global decay factor $(1 + \xi \cdot r)^{-0.5}$ in Eq.~\ref{eq:final_lr_per_layer} controls how aggressively the base LR decreases across rounds, setting the pace at which global optimization slows relative to layer-wise adaptation. The sensitivity of $\xi$ is architecture-dependent: deeper models require a gentler decay to allow layer-wise signals to fully differentiate per-layer rates during the critical mid-training phase, while shallower models converge faster and tolerate earlier decay. As shown in Table~\ref{tab:fedA2L_sensitivity}, $\xi = 0.3$ leads to convergence failure on ResNet-18 and ResNet-34, with best accuracy collapsing to 44.90\% and 44.29\% respectively, confirming that aggressive decay suppresses layer-wise adaptation before the per-layer signals become effective. In contrast, reducing $\xi$ to 0.1 restores stable convergence at 177 and 157 rounds on ResNet-18 and ResNet-34. Meanwhile, CNN achieves its lowest convergence in 24 rounds under the same aggressive setting, as its simpler feature hierarchy reaches the target earlier under faster decay. We adopt $\xi = 0.1$ for ResNet architectures and $\xi = 0.3$ for CNN. 

\textbf{Influence of fusion balance ($\beta$)}.
The fusion balance $\beta$ in Eq.~\ref{eq:fusion} is a key operational parameter, as it controls how FedA2L balances local update intensity $\sigma$ and network consensus constraints $\zeta$ when constructing the per-layer LR signal. The extreme cases $\beta = 0$, which relies solely on network consensus constraints $\zeta$, and $\beta = 1$, which relies solely on local update intensity $\sigma$, serve as direct ablations of each individual component. Both single-signal configurations exhibit slower convergence and lower best accuracy than any balanced setting, as evidenced by Table~\ref{tab:fedA2L_sensitivity} and the accuracy trajectories in Fig.~\ref{fig:beta}. The performance degradation observed in both settings indicates that neither signal alone is sufficient for effective layer-wise adaptation. For ResNet-18 on CIFAR-100, the best test accuracy drops from 49.50\% at $\beta = 0.6$ to 49.23\% at $\beta = 0.0$ and 49.10\% at $\beta = 1.0$, demonstrating that $\sigma$ and $\zeta$ provide complementary and non-redundant information whose combination is necessary for full performance. For deeper architectures such as ResNet-34, performance degradation at $\beta = 1.0$ is more pronounced than at $\beta = 0.0$, with the required rounds increasing from 150 to 159 at $\beta = 1.0$, compared to 158 at $\beta = 0.0$. This suggests that network consensus constraints offer more informative layer-wise signals in complex feature hierarchies. FedA2L remains stable across $\beta \in [0.4, 0.8]$, with $\beta = 0.6$ consistently achieving the strongest convergence speed and accuracy across all evaluated architectures. We therefore adopt $\beta = 0.6$ as the default, reflecting a moderate emphasis on local update intensity while retaining sufficient weight on 
network consensus constraints. 

\begin{table}
\centering
\caption{FedA2L default configuration.}
\label{tab:fedA2L_defaults}
\begin{tabularx}{\columnwidth}{c|l|c|c}
\toprule
Parameter & Tested range & CNN & \makecell{ResNet-18 \\ ResNet-34} \\

\midrule
$R_{\text{warm}}$ & $\{10, 20, 40\}$       & 10  & 40 \\
$\rho$            & $\{5, 10, 20\}$        & 5   & 10 \\
$\tau$            & $\{1e-2, 1e-3, 1e-4\}$   & 1e-3 & 1e-3 \\
$\beta$           & \makecell{$\{0.0, 0.2, 0.4, 0.6,0.8,1.0\}$} & 0.6 & 0.6 \\
$\xi$             & $\{0.05, 0.1, 0.3\}$   & 0.3 & 0.1$^{\dagger}$ \\
\bottomrule
\end{tabularx}
\footnotesize{$^{\dagger}$ $\xi = 0.05$ is applied for TinyImageNet due to higher class diversity.}
\end{table}

\textbf{Overall robustness and recommended configuration}. The default configuration is summarized in Table~\ref{tab:fedA2L_defaults}. Among the five parameters, $\rho$ and $\tau$ show limited sensitivity across all tested architectures, while $\beta$ remains stable within $[0.4, 0.8]$, confirming that the dual-signal fusion is robust to reasonable variation in weighting. The primary configuration parameters are $R_{\text{warm}}$ and $\xi$, both of which interact directly with model depth and should be adjusted when deploying on architectures outside the configurations evaluated here.

\section{Discussion} \label{sec:discuss}

FedA2L dynamically adjusts LRs at the layer level using only locally available model statistics, integrating seamlessly into any DFL protocol without modifying the core aggregation logic or introducing additional communication overhead. This distinguishes it from server-dependent adaptive methods such as FedYogi, layer-wise adaptive approaches that rely on centralized gradient statistics, and divergence-based approaches \citep{hu2024fedmmd,jang2024personalized,yao2025fedrda} that utilize deviation signals for representation alignment or aggregation reweighting. Such a mechanism is particularly relevant under non-IID data and dynamic network conditions where consistent global statistics are difficult to obtain. The layer-wise LR signals in FedA2L may also be extended to personalized DFL, where client-specific adaptation is incorporated into decentralized optimization.

While FedA2L offers flexibility, it involves several practical considerations. First, some hyperparameters (e.g., $R_{\text{warm}}$ and $\xi$) depend on model depth and may require additional tuning across different architectures. Second, the performance gains are less pronounced when the base DFL algorithm already incorporates strong drift correction or adaptive regularization (e.g., FedNTD and FedYogi), where scheduler-based baselines remain competitive. Overall, despite these considerations, FedA2L provides a robust and effective foundation for layer-wise adaptation in decentralized learning.

\section{Conclusion} \label{sec:con}

This paper presents FedA2L, a layer-wise adaptive LR method designed specifically for DFL. By exploiting dual state-transition signals, FedA2L enables per-layer LR modulation that captures layer-specific heterogeneity while stabilizing network consensus, without requiring modification to the underlying DFL protocol. Extensive experiments demonstrate that FedA2L consistently accelerates convergence and improves model accuracy across diverse models, datasets, and DFL algorithms, without additional communication overhead. The reduction in communication rounds further improves communication efficiency, which is beneficial in bandwidth-constrained and latency-sensitive environments. These results indicate that layer-wise adaptation provides an effective mechanism for addressing heterogeneity in decentralized learning, establishing FedA2L as a practical and scalable optimization approach for DFL systems.

\section*{Funding}
    This research was supported by the Basic Science Research Program through the National Research Foundation of Korea (NRF) funded by the Ministry of Education (No. NRF-2022R1I1A3072355).
\printcredits

\bibliographystyle{unsrtnat}
\bibliography{cite}

@article{yosinski2014transferable,
  title={How transferable are features in deep neural networks?},
  author={Yosinski, Jason and Clune, Jeff and Bengio, Yoshua and Lipson, Hod},
  journal={Advances in neural information processing systems},
  volume={27},
  year={2014}
}

@inproceedings{zeiler2014visualizing,
  title={Visualizing and understanding convolutional networks},
  author={Zeiler, Matthew D and Fergus, Rob},
  booktitle={Computer Vision--ECCV 2014: 13th European Conference, Zurich, Switzerland, September 6-12, 2014, Proceedings, Part I 13},
  pages={818--833},
  year={2014},
  organization={Springer}
}

@inproceedings{ro2021autolr,
  title={{AutoLR}: Layer-wise pruning and auto-tuning of learning rates in fine-tuning of deep networks},
  author={Ro, Youngmin and Choi, Jin Young},
  booktitle={Proceedings of the AAAI Conference on Artificial Intelligence},
  volume={35},
  pages={2486--2494},
  year={2021}
}

@article{li2020federated,
  title={Federated optimization in heterogeneous networks},
  author={Li, Tian and Sahu, Anit Kumar and Zaheer, Manzil and Sanjabi, Maziar and Talwalkar, Ameet and Smith, Virginia},
  journal={Proceedings of Machine learning and systems},
  volume={2},
  pages={429--450},
  year={2020}
}

@article{reddi2021adaptive,
  title={Adaptive federated optimization},
  author={Reddi, Sashank and Charles, Zachary and Zaheer, Manzil and Garrett, Zachary and Rush, Keith and Kone{\v{c}}n{\`y}, Jakub and Kumar, Sanjiv and McMahan, H Brendan},
  journal={arXiv preprint arXiv:2003.00295},
  year={2021}
}

@article{lee2021fedntd,
  author       = {Gihun Lee and
                  Yongjin Shin and
                  Minchan Jeong and
                  Se{-}Young Yun},
  title        = {Preservation of the Global Knowledge by Not-True Self Knowledge Distillation
                  in Federated Learning},
  journal      = {CoRR},
  volume       = {abs/2106.03097},
  year         = {2021},
  eprinttype    = {arXiv},
  eprint       = {2106.03097},
  bibsource    = {dblp computer science bibliography, https://dblp.org}
}

@Techreport{krizhevsky2009learning,
 author = {Krizhevsky, Alex and Hinton, Geoffrey},
 address = {Toronto, Ontario},
 institution = {University of Toronto},
 number = {0},
 publisher = {Technical report, University of Toronto},
 title = {Learning multiple layers of features from tiny images},
 year = {2009},
 title_with_no_special_chars = {Learning multiple layers of features from tiny images},
}

@misc{kim2025hyperboliclrepochinsensitivelearning,
      title={{HyperbolicLR}: Epoch insensitive learning rate scheduler}, 
      author={Tae-Geun Kim},
      year={2025},
      eprint={2407.15200},
      archivePrefix={arXiv},
      primaryClass={cs.LG},
      url={https://arxiv.org/abs/2407.15200}, 
}

@article{zhao2018federated,
  title={Federated Learning with {non-IID} Data},
  author={Zhao, Yue and Li, Meng and Lai, Liangzhen and Suda, Naveen and Civin, Damon and Chandra, Vikas},
  journal={arXiv preprint arXiv:1806.00582},
  year={2018}
}

@inproceedings{zheng2024federated,
  title={Federated Learning via Consensus Mechanism on Heterogeneous Data: A New Perspective on Convergence},
  author={Zheng, Shu and Ye, Tiandi and Li, Xiang and Gao, Ming},
  booktitle={ICASSP 2024-2024 IEEE International Conference on Acoustics, Speech and Signal Processing (ICASSP)},
  pages={7595--7599},
  year={2024},
  organization={IEEE}
}

@inproceedings{lalitha2018fully,
  title={Fully decentralized federated learning},
  author={Lalitha, Anusha and Shekhar, Shubhanshu and Javidi, Tara and Koushanfar, Farinaz},
  booktitle={Third workshop on bayesian deep learning (NeurIPS)},
  volume={12},
  year={2018}
}

@inproceedings{mcmahan2017communication,
  title={Communication-efficient learning of deep networks from decentralized data},
  author={McMahan, Brendan and Moore, Eider and Ramage, Daniel and Hampson, Seth and y Arcas, Blaise Aguera},
  booktitle={Artificial intelligence and statistics},
  pages={1273--1282},
  year={2017},
  organization={PMLR}
}

@article{yuan2024decentralized,
  title={Decentralized Federated Learning: A Survey and Perspective},
  author={Yuan, Liangqi and Wang, Ziran and Sun, Lichao and Yu, Philip S. and Brinton, Christopher G.},
  journal={IEEE Internet of Things Journal},
  volume={11},
  number={21},
  pages={34617--34638},
  year={2024},
  doi={10.1109/JIOT.2024.3407584},
}

@article{lian2017can,
  title={Can decentralized algorithms outperform centralized algorithms? a case study for decentralized parallel stochastic gradient descent},
  author={Lian, Xiangru and Zhang, Ce and Zhang, Huan and Hsieh, Cho-Jui and Zhang, Wei and Liu, Ji},
  journal={Advances in neural information processing systems},
  volume={30},
  year={2017}
}

@inproceedings{shi2025fedawa,
  title={{FedAWA}: Adaptive Optimization of Aggregation Weights in Federated Learning Using Client Vectors},
  author={Shi, Changlong and Zhao, He and Zhang, Bingjie and Zhou, Mingyuan and Guo, Dandan and Chang, Yi},
  booktitle={Proceedings of the Computer Vision and Pattern Recognition Conference},
  pages={30651--30660},
  year={2025}
}

@article{DFedHPO,
title = {Consensus-Driven Hyperparameter Optimization for Accelerated Model Convergence in Decentralized Federated Learning},
journal = {Internet of Things},
volume = {30},
pages = {101476},
year = {2025},
issn = {2542-6605},
doi = {https://doi.org/10.1016/j.iot.2024.101476},
author = {Anam Nawaz Khan and Qazi Waqas Khan and Atif Rizwan and Rashid Ahmad and Do Hyeun Kim},
}

@inproceedings{tang2024adapted,
  title={Adapted weighted aggregation in federated learning},
  author={Tang, Yitong},
  booktitle={Proceedings of the AAAI Conference on Artificial Intelligence},
  volume={38},
  pages={23763--23765},
  year={2024}
}

@inproceedings{Koloskova2019,
  author    = {Koloskova, Anastasia and Stich, Sebastian U. and Jaggi, Martin},
  title     = {Decentralized Stochastic Optimization and Gossip Algorithms with Compressed Communication},
  booktitle = {Proceedings of the 36th International Conference on Machine Learning (ICML)},
  series    = {Proceedings of Machine Learning Research},
  volume    = {97},
  pages     = {3478--3487},
  year      = {2019}
}

@article{tang2022gossipfl,
  title={{GossipFL}: A decentralized federated learning framework with sparsified and adaptive communication},
  author={Tang, Zhenheng and Shi, Shaohuai and Li, Bo and Chu, Xiaowen},
  journal={IEEE Transactions on Parallel and Distributed Systems},
  volume={34},
  number={3},
  pages={909--922},
  year={2022},
  publisher={IEEE}
}

@article{nedic2018network,
  title={Network topology and communication-computation tradeoffs in decentralized optimization},
  author={Nedi{\'c}, Angelia and Olshevsky, Alex and Rabbat, Michael G},
  journal={Proceedings of the IEEE},
  volume={106},
  number={5},
  pages={953--976},
  year={2018},
  publisher={IEEE}
}

@ARTICLE{van2025performance,
  author={van Truong, Vo and Khanh Quan, Pham and Park, Dong-Hwan and Kim, Taehong},
  journal={IEEE Access}, 
  title={Performance Evaluation of Decentralized Federated Learning: Impact of Fully and K-Connected Topologies, Heterogeneous Computing Resources, and Communication Bandwidth}, 
  year={2025},
  volume={13},
  number={},
  pages={32741-32755},
  doi={10.1109/ACCESS.2025.3542772}}

@misc{elhussein2025playerfl,
      title={{PLayer-FL}: A Principled Approach to Personalized Layer-wise Cross-Silo Federated Learning}, 
      author={Ahmed Elhussein and Gamze Gürsoy},
      year={2025},
      eprint={2502.08829},
      archivePrefix={arXiv},
      primaryClass={cs.LG},
      url={https://arxiv.org/abs/2502.08829}, 
}

@article{wang2023distributed,
  title={Distributed and secure federated learning for wireless computing power networks},
  author={Wang, Peng and Sun, Wen and Zhang, Haibin and Ma, Wenqiang and Zhang, Yan},
  journal={IEEE Transactions on Vehicular Technology},
  volume={72},
  number={7},
  pages={9381--9393},
  year={2023},
  publisher={IEEE}
}

@article{wang2021edge,
  title={Edge-based communication optimization for distributed federated learning},
  author={Wang, Tian and Liu, Yan and Zheng, Xi and Dai, Hong-Ning and Jia, Weijia and Xie, Mande},
  journal={IEEE Transactions on Network Science and Engineering},
  volume={9},
  number={4},
  pages={2015--2024},
  year={2021},
  publisher={IEEE}
}

@ARTICLE{wang2025verifydfl,
  author={Wang, Shuai and Tian, Youliang and Xiong, Jinbo and Ma, Jianfeng and Zhang, Yan},
  journal={IEEE Transactions on Cognitive Communications and Networking}, 
  title={{VerifyDFL}: Secure Aggregation for Decentralized Federated Learning With Input Validation in Mobile Edge Intelligence}, 
  year={2026},
  volume={12},
  number={},
  pages={2526-2541},
  doi={10.1109/TCCN.2025.3587771}}

@ARTICLE{11134848,
  author={Liu, Jianchun and Yan, Jiaming and Xu, Hongli and Wang, Lun and Wang, Zhiyuan and Huang, Jinyang and Qiao, Chunming},
  journal={IEEE Transactions on Networking}, 
  title={Accelerating Decentralized Federated Learning With Probabilistic Communication in Heterogeneous Edge Computing}, 
  year={2026},
  volume={34},
  number={},
  pages={486-501},
  doi={10.1109/TON.2025.3600015}}

@inproceedings{gao2022feddc,
  title={{FedDC}: Federated Learning with {non-IID} Data via Local Drift Decoupling and Correction},
  author={Gao, Liang and Fu, Huazhu and Li, Li and Chen, Yingwen and Xu, Ming and Xu, Cheng-Zhong},
  booktitle={Proceedings of the IEEE/CVF conference on computer vision and pattern recognition},
  pages={10112--10121},
  year={2022}
}

@inproceedings{chen2024optimizing,
  title     = {Optimizing Personalized Federated Learning Through Adaptive Layer-Wise Learning},
  author    = {Chen, Weihang and Yang, Cheng and Ren, Jie and Li, Zhiqiang and Wang, Zheng},
  booktitle = {Proceedings of the Thirty-Fourth International Joint Conference on
               Artificial Intelligence, {IJCAI-25}},
  publisher = {International Joint Conferences on Artificial Intelligence Organization},
  editor    = {James Kwok},
  pages     = {4860--4868},
  year      = {2025},
  month     = {August},
  note      = {Main Track},
  doi       = {10.24963/ijcai.2025/541},
}

@article{beltran2023decentralized,
  title={Decentralized federated learning: Fundamentals, state of the art, frameworks, trends, and challenges},
  author={Beltr{\'a}n, Enrique Tom{\'a}s Mart{\'\i}nez and P{\'e}rez, Mario Quiles and S{\'a}nchez, Pedro Miguel S{\'a}nchez and Bernal, Sergio L{\'o}pez and Bovet, G{\'e}r{\^o}me and P{\'e}rez, Manuel Gil and P{\'e}rez, Gregorio Mart{\'\i}nez and Celdr{\'a}n, Alberto Huertas},
  journal={IEEE Communications Surveys \& Tutorials},
  volume={25},
  number={4},
  pages={2983--3013},
  year={2023},
  publisher={IEEE}
}

@article{xiao2025flare,
  title={{FLARE}: A New Federated Learning Framework with Adjustable Learning Rates over Resource-Constrained Wireless Networks},
  author={Xiao, Bingnan and Zhang, Jingjing and Ni, Wei and Wang, Xin},
  journal={IEEE Transactions on Wireless Communications},
  year={2025},
  publisher={IEEE}
}

@article{kundroo2023federated,
  title={Federated learning with hyper-parameter optimization},
  author={Kundroo, Majid and Kim, Taehong},
  journal={Journal of King Saud University-Computer and Information Sciences},
  volume={35},
  number={9},
  pages={101740},
  year={2023},
  publisher={Elsevier}
}

@inproceedings{nakka2024federated,
  title={Federated hyperparameter optimization through reward-based strategies: Challenges and insights},
  author={Nakka, Krishna Kanth and Frikha, Ahmed and Mendis, Ricardo and Jiang, Xue and Zhou, Xuebing},
  booktitle={Proceedings of the IEEE/CVF Conference on Computer Vision and Pattern Recognition},
  pages={4236--4244},
  year={2024}
}

@inproceedings{karimi2023fed,
  title={{Fed-LAMB}: layer-wise and dimension-wise locally adaptive federated learning},
  author={Karimi, Belhal and Li, Ping and Li, Xiaoyun},
  booktitle={Uncertainty in Artificial Intelligence},
  pages={1037--1046},
  year={2023},
  organization={PMLR}
}

@article{shi2025fedlws,
  title={{FedLWS}: Federated Learning with Adaptive Layer-wise Weight Shrinking},
  author={Shi, Changlong and Li, Jinmeng and Zhao, He and Guo, Dandan and Chang, Yi},
  journal={arXiv preprint arXiv:2503.15111},
  year={2025}
}

@article{le2015tiny,
  title={Tiny imagenet visual recognition challenge},
  author={Le, Yann and Yang, Xuan},
  journal={CS 231N},
  volume={7},
  number={7},
  pages={3},
  year={2015}
}

@misc{loshchilov2017sgdrstochasticgradientdescent,
      title={{SGDR}: Stochastic Gradient Descent with Warm Restarts}, 
      author={Ilya Loshchilov and Frank Hutter},
      year={2017},
      eprint={1608.03983},
      archivePrefix={arXiv},
      primaryClass={cs.LG},
      url={https://arxiv.org/abs/1608.03983}, 
}

@misc{smith2018superconvergencefasttrainingneural,
      title={Super-Convergence: Very Fast Training of Neural Networks Using Large Learning Rates}, 
      author={Leslie N. Smith and Nicholay Topin},
      year={2018},
      eprint={1708.07120},
      archivePrefix={arXiv},
      primaryClass={cs.LG},
      url={https://arxiv.org/abs/1708.07120}, 
}

@article{ZHOU2024120582,
title = {{DeFTA}: A plug-and-play peer-to-peer decentralized federated learning framework},
journal = {Information Sciences},
volume = {670},
pages = {120582},
year = {2024},
issn = {0020-0255},
doi = {https://doi.org/10.1016/j.ins.2024.120582},

author = {Yuhao Zhou and Minjia Shi and Yuxin Tian and Qing Ye and Jiancheng Lv},
}

@article{WANG2023449,
title = {Enhancing privacy preservation and trustworthiness for decentralized federated learning},
journal = {Information Sciences},
volume = {628},
pages = {449-468},
year = {2023},
issn = {0020-0255},
doi = {https://doi.org/10.1016/j.ins.2023.01.130},
author = {Lingling Wang and Xueqin Zhao and Zhongkai Lu and Lin Wang and Shouxun Zhang}
}

@article{molo2025decentralized,
title = {Decentralized edge learning: A comparative study of distillation strategies and dissimilarity measures},
journal = {Future Generation Computer Systems},
volume = {176},
pages = {108171},
year = {2026},
issn = {0167-739X},
publisher={Elsevier},
doi = {https://doi.org/10.1016/j.future.2025.108171},
author = {Mbasa Joaquim Molo and Lucia Vadicamo and Claudio Gennaro and Emanuele Carlini},
}

@article{selo2025fedtvd,
  title={{FedTVD}: Balancing Data Quality and Quantity for Robust Federated Learning},
  author={Selo, Radwan and Kundroo, Majid and Kim, Taehong},
  journal={Future Generation Computer Systems},
  pages={108177},
  year={2025},
  publisher={Elsevier}
}

@article{park2025def,
title = {{Def-Ag}: An energy-efficient decentralized federated learning framework via aggregator clients},
journal = {Future Generation Computer Systems},
volume = {175},
pages = {108114},
year = {2026},
issn = {0167-739X},
doi = {https://doi.org/10.1016/j.future.2025.108114},
author = {Junyoung Park and Sungpil Woo and Joohyung Lee},
}

@ARTICLE{9244624,
  author={Sun, Wen and Lei, Shiyu and Wang, Lu and Liu, Zhiqiang and Zhang, Yan},
  journal={IEEE Transactions on Industrial Informatics}, 
  title={Adaptive Federated Learning and Digital Twin for Industrial Internet of Things}, 
  year={2021},
  volume={17},
  number={8},
  pages={5605-5614},
  doi={10.1109/TII.2020.3034674}}

@article{padmavathi2025digital,
  title={Digital twin driven smart factories: real time physics based co-simulation using edge ai and federated learning},
  author={Padmavathi, V and Kanimozhi, R and Saminathan, R},
  journal={Scientific Reports},
  volume={15},
  number={1},
  pages={43373},
  year={2025},
  publisher={Nature Publishing Group UK London}
}

@inproceedings{10.5555/3692070.3693383,
author = {Liu, Yong and Zhang, Haoran and Li, Chenyu and Huang, Xiangdong and Wang, Jianmin and Long, Mingsheng},
title = {Timer: generative pre-trained transformers are large time series models},
year = {2024},
publisher = {JMLR.org},
booktitle = {Proceedings of the 41st International Conference on Machine Learning},
articleno = {1313},
numpages = {31},
location = {Vienna, Austria},
series = {ICML'24}
}

@InProceedings{DFedSAM,
  title = 	 {Improving the Model Consistency of Decentralized Federated Learning},
  author =       {Shi, Yifan and Shen, Li and Wei, Kang and Sun, Yan and Yuan, Bo and Wang, Xueqian and Tao, Dacheng},
  booktitle = 	 {Proceedings of the 40th International Conference on Machine Learning},
  pages = 	 {31269--31291},
  year = 	 {2023},
  volume = 	 {202},
  series = 	 {Proceedings of Machine Learning Research},
  month = 	 {23--29 Jul},
  publisher =    {PMLR}
}

@article{zhang2017cautionary,
  title={Cautionary tales on air-quality improvement in Beijing},
  author={Zhang, Shuyi and Guo, Bin and Dong, Anlan and He, Jing and Xu, Ziping and Chen, Song Xi},
  journal={Proceedings of the Royal Society A: Mathematical, Physical and Engineering Sciences},
  volume={473},
  number={2205},
  year={2017},
  publisher={The Royal Society}
}

@inproceedings{lai2018modeling,
  title={Modeling long-and short-term temporal patterns with deep neural networks},
  author={Lai, Guokun and Chang, Wei-Cheng and Yang, Yiming and Liu, Hanxiao},
  booktitle={The 41st international ACM SIGIR conference on research \& development in information retrieval},
  pages={95--104},
  year={2018}
}

@article{chen2024resilient,
  title={Resilient collaborative caching for multi-edge systems with robust federated deep learning},
  author={Chen, Zheyi and Liang, Jie and Yu, Zhengxin and Cheng, Hongju and Min, Geyong and Li, Jie},
  journal={IEEE Transactions on Networking},
  volume={33},
  number={2},
  pages={654--669},
  year={2024},
  publisher={IEEE}
}

@ARTICLE{10839118,
  author={Chen, Zheyi and Jiang, Qingnan and Chen, Lixian and Chen, Xing and Li, Jie and Min, Geyong},
  journal={IEEE Transactions on Mobile Computing}, 
  title={{MC-2PF}: A Multi-Edge Cooperative Universal Framework for Load Prediction With Personalized Federated Deep Learning}, 
  year={2025},
  volume={24},
  number={6},
  pages={5138-5154},
  doi={10.1109/TMC.2025.3528404}}

@ARTICLE{10704927,
  author={Chen, Zheyi and Zhang, Junjie and Min, Geyong and Ning, Zhaolong and Li, Jie},
  journal={IEEE Journal on Selected Areas in Communications}, 
  title={{Traffic-Aware Lightweight Hierarchical Offloading Toward Adaptive Slicing-Enabled SAGIN}}, 
  year={2024},
  volume={42},
  number={12},
  pages={3536-3550},
  doi={10.1109/JSAC.2024.3459020}}

@article{yao2025fedrda,
  title={{FedRDA: Representation Deviation Alignment in Heterogeneous Federated Learning}},
  author={Yao, Wenjie and Sun, Guanglu and Zhu, Suxia and Wang, Ruidong and Zhu, Xinzhong and Xu, HuiYing and Wei, Xiguang},
  journal={IEEE Transactions on Industrial Informatics},
  year={2025},
  publisher={IEEE}
}

@inproceedings{jang2024personalized,
  title={Personalized federated learning via deviation tracking representation learning},
  author={Jang, Jaewon and Choi, Bong Jun},
  booktitle={2024 International Conference on Information Networking (ICOIN)},
  pages={762--766},
  year={2024},
  organization={IEEE}
}

@article{hu2024fedmmd,
  title={{FedMMD: A Federated weighting algorithm considering Non-IID and Local Model Deviation}},
  author={Hu, Kai and Li, Yaogen and Zhang, Shuai and Wu, Jiasheng and Gong, Sheng and Jiang, Shanshan and Weng, Liguo},
  journal={Expert Systems with Applications},
  volume={237},
  pages={121463},
  year={2024},
  publisher={Elsevier}
}

\end{document}